\documentclass[acmsmall]{acmart}

\AtBeginDocument{%
  \providecommand\BibTeX{{%
    \normalfont B\kern-0.5em{\scshape i\kern-0.25em b}\kern-0.8em\TeX}}}

\setcopyright{acmcopyright}
\copyrightyear{2021}
\acmYear{2021}
\acmDOI{}

\usepackage{amsmath}
\usepackage{amsfonts} 
\usepackage{balance}
\usepackage{xcolor}
\usepackage{caption,subcaption}
\usepackage{arydshln}
\usepackage{amsthm}
\usepackage{longtable}
\newtheorem{definition}{Definition}
\usepackage{graphicx}
\usepackage{marvosym}
\usepackage{draftwatermark}
\usepackage{longtable}

\usepackage{lscape}
\usepackage{rotating}
\usepackage{amsmath}
\usepackage{multicol}
\usepackage{color}
\usepackage{multirow}

\usepackage{pifont}
\usepackage[english]{babel}
\usepackage[utf8]{inputenc}
\usepackage{algorithm}
\usepackage{xcolor}
\usepackage{tikz}

\usepackage{blindtext}
\usepackage{hyperref}
\usepackage{longtable}
\usepackage{pbox}
\usepackage{graphicx}
\usepackage{multicol}
\usepackage{multirow}
\usepackage{balance}

\usepackage{natbib}
\usepackage{multirow}
\usepackage{multicol}
\usepackage{url}

\usepackage{threeparttable}

\usepackage{adjustbox}
\usepackage{titlesec}

\usepackage{color, colortbl}
\definecolor{beige}{rgb}{0.96, 0.96, 0.86}
 \definecolor{blanchedalmond}{rgb}{1.0, 0.92, 0.8}
\definecolor{bubbles}{rgb}{0.91, 1.0, 1.0}
\definecolor{eggshell}{rgb}{0.94, 0.92, 0.84}
\definecolor{ghostwhite}{rgb}{0.97, 0.97, 1.0}
\begin{document}

\title{Open-world Machine Learning: Applications, Challenges, and Opportunities}

\SetWatermarkText{Preprint}
\SetWatermarkScale{0.3}
\author{Jitendra Parmar}
\email{2019rcp9044@mnit.ac.in }
\orcid{}
\author{Satyendra S. Chouhan}
\email{sschouhan@mnit.ac.in}
\affiliation{%
  \institution{Malaviya National Institute of Technology Jaipur}
  \streetaddress{JLN Marg}
  \city{Jaipur}
  \state{Rajsthan}
  \country{INDIA}
  \postcode{302017}
}

\author{Vaskar Raychoudhury}
\affiliation{%
  \institution{Miami University}
  \streetaddress{510 E. High St.}
  \city{Oxford, Ohio}
  \country{USA}}
\email{raychov@miamioh.edu}

\author{Santosh S. Rathore}
\affiliation{%
 \institution{ABV-IIITM Gwalior}
 \streetaddress{}
 \city{Gwalior}
 \state{MadhyaPradesh}
 \country{INDIA}}

\begin{abstract}

Traditional machine learning mainly supervised learning, follows the assumptions of closed-world learning, i.e., for each testing class, a training class is available. However, such machine learning models fail to identify the classes which were not available during training time. These classes can be referred to as unseen classes. Whereas open-world machine learning (OWML) deals with unseen classes. In this paper, first, we present an overview of OWML with importance to the real-world context. Next, different dimensions of open-world machine learning are explored and discussed. The area of OWML gained the attention of the research community in the last decade only. We have searched through different online digital libraries and scrutinized the work done in the last decade. This paper presents a systematic review of various techniques for OWML. It also presents the research gaps, challenges, and future directions in open-world machine learning. This paper will help researchers understand the comprehensive developments of OWML and the likelihood of extending the research in suitable areas. It will also help to select applicable methodologies and datasets to explore this further.

\end{abstract}

\ccsdesc[500]{Computing methodologies~Learning paradigms}

\keywords{Open-world Learning, Continual Machine Learning, Incremental Learning, Open-world Image and Text Classification}

\maketitle

\section{Introduction}

\begin{figure}[htb]
\begin{subfigure}{.4\textwidth}
  \includegraphics[height = 1.3 in, width = 2.2 in]{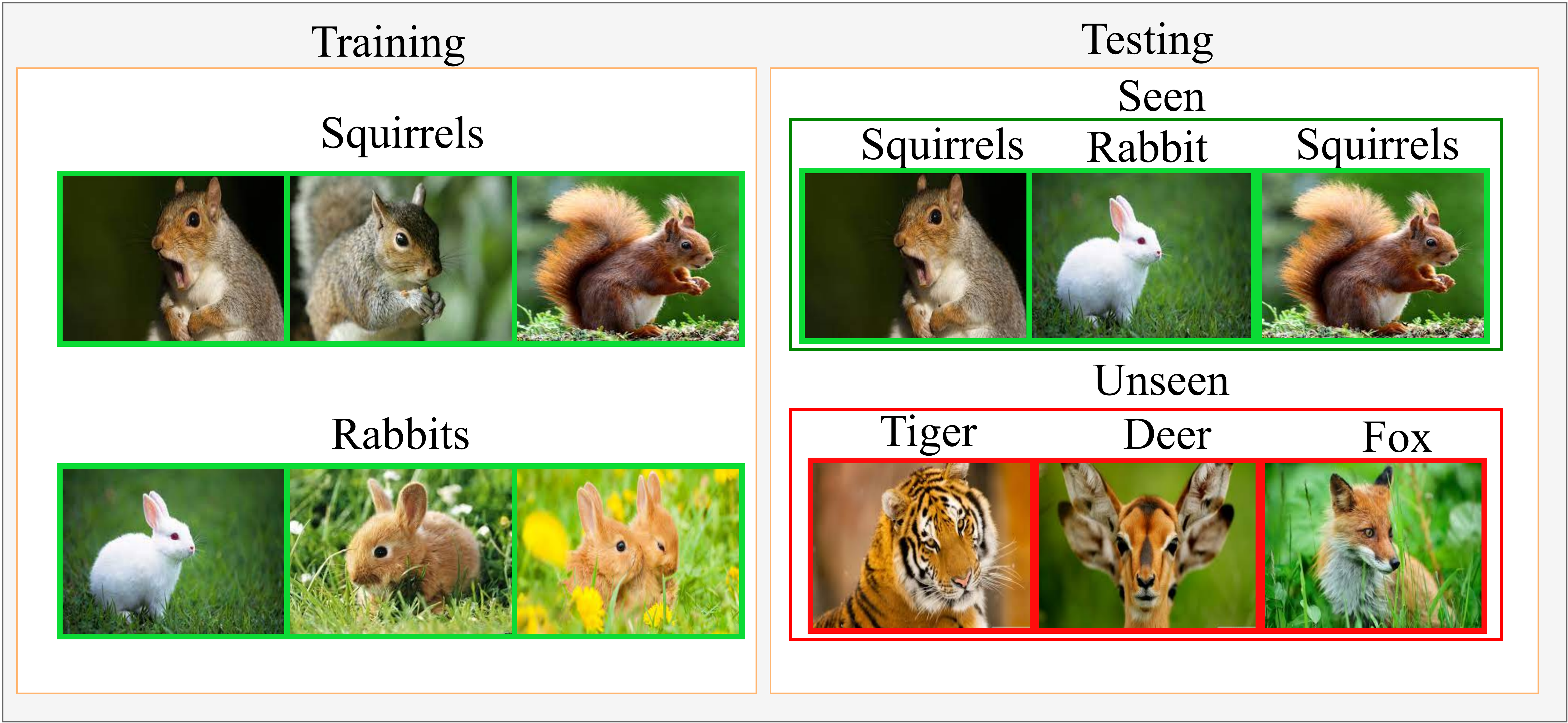}  
  \caption{Image Classification with Open-world Machine Learning}
  \label{OWMLinCVIP}
\end{subfigure}
\begin{subfigure}{.4\textwidth}
  \includegraphics[height = 1.4 in, width = 2.6 in]{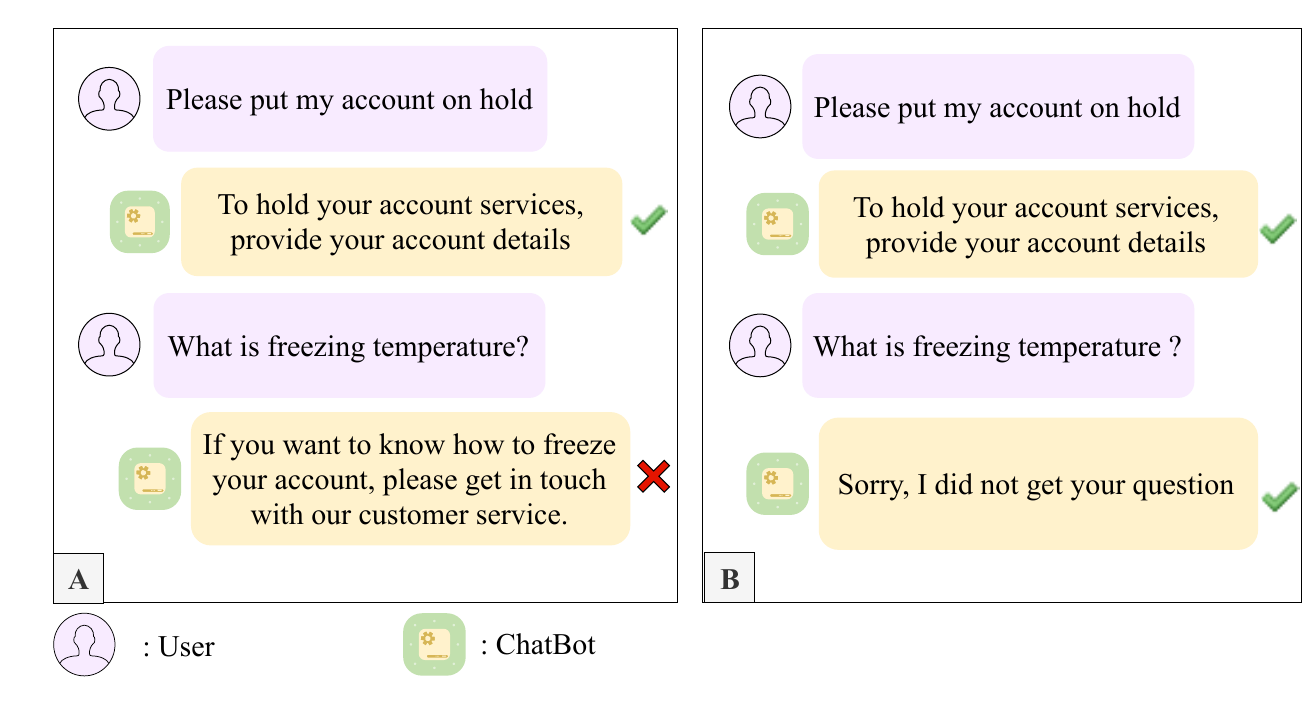}  
  \caption{ Message Exchange between the User and \textit{abc} Banking ChatBot}
  \label{OWMLinNLP}
\end{subfigure}
\caption{(a) OWML in Computer Vision and Image Processing (CV-IP) and (b) OWML in Natural Language Processing (NLP)}
\label{OWMLExample}
\end{figure}

Traditional machine learning approaches have promising outcomes for every domain of data analysis. For the last several years, it has played an essential role in many data fields, which is effective in many domains, so it is a progressing method for data analytic and visualization. However, traditional machine learning has some limitations, such as 1) it works with isolated data and learns without using previous knowledge, and 2) a trained machine model can only work with the input instances for which similar instances have been used for the training purposes~\cite{kotsiantis2007supervised}. Consider the example  which illustrated OWML in Computer Vision and Image Processing  (CV-IP) and Natural Language Processing (NLP) field (Figure~\ref{OWMLExample}).

Assume that a machine learning model is trained to identify the specific images, say Squirrels and Rabbits. The complete training of this model is done with the different images of Squirrels and Rabbits. If we provide either Squirrels or Rabbits at the testing time, it is competent to identify Squirrels and Rabbits. If we provide an image other than a Squirrel or Rabbits, it will still classify it as Squirrels or Rabbits.

However, in the case of OWML, the images of Tiger, Deer, and Fox will be rejected at the time of testing, as it is unseen for the model, i.e., 
the system can detect examples that are not from the training set. The capability to identify examples as unseen or classify them is called open-world learning.

In Figure (\ref{OWMLinNLP}), a banking ChatBot is specifically designed for account and transaction-related queries. What will happen if the user asks an out-of-scope question? The system will reply with a false response since it is trained in a closed-world environment (Scenario A). 
In the first inquiry, the user asks about the "put services of account on hold." The query is correctly identified by the ChatBot and responds appropriately. In the second inquiry user ask about the freezing point of water, which is an out-of-scope query. However, ChatBot identifies it incorrectly and suggests the procedure of "account freezing." 
Scenario B  presents an ideal ChatBot system using open-world machine learning. Both the inquiries are the same as Scenario A, but the ChatBot correctly identifies the second query's intent. The intent of the second query is out-of-scope; hence ChatBot refuge to answer this query.

Classical machine learning~\cite{michie1994machine,alpaydin2020introduction,svensen2007pattern,murphy2012machine}, especially supervised learning~\cite{kotsiantis2007supervised}, follows the assumptions of close-world learning~\cite{li2017learning}. Where for each testing class, a training class is available~\cite{moore2015context,michie1994machine}. However,  in a real-world scenario, interactive and automated applications work in a dynamic environment, and data from new classes arrive regularly.
In such cases, the model that follows closed-world assumptions cannot address that kind of situation. Open-world learning is a universal phenomenon, i.e., it is not limited to specific machine learning~\cite{murphy2012machine}. It can be generally defined as a model that can learn new things that are not learned before while performing its anticipated job. Learning new things during work is a capability to identify the unknown. It can fill the learning style gap between human beings and machines. In other words, we can say that open-world machine learningcan enable the machine to learn as a normal human being does. It is self-motivated learning that can distinguish whenever something different appears. Open-world machine learning can enhance several recent AI-based prototypes in CV-IP such as self-driving cars~\cite{bojarski2016end,yang2014vehicle}, healthcare and medical diagnosis~\cite{coatrieux2006review,razzak2018deep}, video surveillance~\cite{inacio2021osvidcap}, robotics~\cite{stevens2021efficient}, recognition of disruptive images on social media~\cite{jeya2021content,khan2021ai}. Similarly, in NLP, open-world machine learningcan help to improve ChatBot systems~\cite{lokman2018modern,denecke2021artificial}, intelligent assistants~\cite{jiang2015automatic,cowan2017can},email spam detection~\cite{wang2010detecting}, product recommendation~\cite{zhao2016exploring,xiao2007commerce}, and cyberbullying identification~\cite{paul2020cyberbert}.

Open-world learning is a relatively new domain in machine learning, and to the best of our knowledge, there are very few review articles that discuss open-world machine learning. Most of the existing review articles of open-world machine learning are task or domain-specific, such as, In~\cite{shu2017doc}, authors reviewed numerous methodologies that can find novel attacks or malware in the open world. In~\cite{leng2019survey,ye2021deep}, authors reviewed numerous methodologies, including deep learning-based approaches, which were used to identify human beings (person) in an open-world environment. One of the recent survey articles, given in~\cite{geng2020recent}, on open-world machine learning, has reviewed most of the work done in computer vision and image processing. However, in recent years, open-world machine learning has also been used in natural language processing, such as automated dialog-based systems. To the best of our knowledge, there is a lack of review articles available for open-world machine learning, which can provide a broader classification of open-world learning.

This paper presents a systematic review of related works in open-world learning. First, we present an overview of open-world machine learning with importance to the real-world context. It also presents a taxonomic classification of numerous open-world machine learning methods used in computer vision and natural language processing. In addition, we have presented the tabular summaries of existing works, emphasizing the discussion of advantages and disadvantages.  In addition, we discussed some of the baseline benchmark algorithms used in open-world machine learning for both computer vision and natural language processing. The summarised discussion can be helpful in the selection of appropriate methods for a particular problem in a given learning environment. In summary, the contributions of the paper are as follows.
\begin{itemize}
    \item We present a task-based taxonomy that distinguishes Open-world Machine Learning (OWML) key features and their relationships.
    \item We analyzed several techniques and their features in terms of efficiency and other parameters.
    \item We also discussed various datasets and their characteristics used in OWML for computer vision and image processing, and natural language processing to thoroughly understand the outcomes. 
    \item We present various research gaps and challenges in brief, helping to extend the work towards the OWML.
    \item Further, We present some of the associated fields of OWML to determine open-world problems through different techniques.
\end{itemize}

The organization of this paper is as follows. Section 2 presents the background information about OWML. Section 3 explains Review methodology adopted for this paper and the taxonomy of OWML. Section 4  addresses related works of OWML in Computer Vision and Image Processing (CV-IP).  Section 5  addresses related works of OWML in Natural Language Processing (NLP). Section 6 reviews the standard benchmark datasets that several researchers used in the OWML. Section 7 discusses some of the baseline algorithms used in OWML. Section 8 discusses related areas to OWML. Next, we explain some of the research challenges and future directions in OWML (Section 9 and 10). Finally, section 10 concludes this paper.

\section{Background and Formal Definition}

Classical machine learning works in two parts: training and testing. For each example of testing, we must have a training example to identify such classes. Therefore, experts always suggest a high score for testing, but the high testing score cannot guarantee meaningful real-world outcomes. For good results in the real world, the machine needs to learn new things like humans. If the machine learns new things,  especially those not present during training, and recognizes those things in testing, then the system will produce more convincing outputs. open-world machine learning can address the concerns of a dynamic environment where the input and nature of input data (size, category, frequency, etc.) are changing rapidly.

To better understand open-world learning, we have to know what \textit{open} means. The systems are often designed for a specific task; the models are trained to identify particular objects if we consider computer vision examples. However, do similar objects come in the real world? In real-world objects are surrounded by many other things. In open-world learning, classifications are open, or models can learn incrementally. It can learn about new classes and update the existing model (without re-training). open-world machine learning also refers as cumulative learning~\cite{alpaydin2020introduction} and open-world recognition~\cite{svensen2007pattern,deng2009imagenet}. Before comparing classical machine learning techniques with open-world machine learning, we have defined some terms here. 
\begin{itemize}
    \item[1.] \textit{Seen-Seen Instances}: Instances that are labelled in the training datasets i.e., classes are know \textit{a priori}.
    \item [2.] \textit{Seen-Unseen Instances}: Instances that are unlabelled in testing datasets but belongs to the seen classes i.e., classes are known during training time.
    \item[3.] \textit{Unseen-Unseen Instances}: Instances that are unlabelled in datasets and have not been appeared during training time.
    \item[4.] \textit{Unseen Instances}: unlabelled instance during training time.
\end{itemize}

\begin{table}[htb]
\scriptsize
\caption{Different Paradigms of Machine Learning}
\label{MlvsCML}
\begin{tabular}{|p{1.0cm}|p{3.2cm}|p{1.2cm}|p{1.5cm}|p{1.7cm}|p{1.4cm}|p{1.4cm}|}
\hline
\rowcolor{blanchedalmond}
\textbf{Domain}                                      & \textbf{Techniques and Proposed Year}                      & \textbf{Task}                          & \textbf{Training Data}        & \textbf{Testing Data}                  & \textbf{Knowledge Accumulation} & \textbf{Knowledge Retention}\\\hline

\multirow{4}{*}{\textbf{ML}} 

& \cellcolor{beige} Supervised Learning (1988)                        &\cellcolor{beige} CL and RG &\cellcolor{beige} Seen-Seen                 &\cellcolor{beige} Seen-Unseen                          &\cellcolor{beige} -                      &\cellcolor{beige} -                   \\ \cline{2-7}
                                           
                                            &\cellcolor{beige} Unsupervised  Learning (1989)                     &\cellcolor{beige} CR and AS   &\cellcolor{beige} Unseen               &\cellcolor{beige} Seen-Unseen       &\cellcolor{beige} -            &\cellcolor{beige} -                   \\ \cline{2-7}
                                            &\cellcolor{beige} Reinforcement Learning (1995)                     &\cellcolor{beige} CL, CR and CNT    &\cellcolor{beige} Seen-Seen /Unseen                 &\cellcolor{beige} Seen-Unseen           &\cellcolor{beige} -                      &\cellcolor{beige} -                   \\ \cline{2-7}
                              &\cellcolor{beige} Semi-Supervised Learning (2000)                   &\cellcolor{beige} CR and CL &\cellcolor{beige} Seen-Seen / Unseen &\cellcolor{beige} Seen-Unseen          &\cellcolor{beige} -                      &\cellcolor{beige} -                   \\  \hline
\textbf{DL}                              &\cellcolor{ghostwhite} Deep Neural Networks (1965)                       & \cellcolor{ghostwhite} CL, CR and RL & \cellcolor{ghostwhite} Seen-Seen / unseen                 &\cellcolor{ghostwhite} Seen-Unseen/ Unseen                          & \cellcolor{ghostwhite} -                      & \cellcolor{ghostwhite} -                   \\ \hline
\multirow{5}{*}{\textbf{CML}} &\cellcolor{bubbles} Supervised Continual Learning (1995)              &\cellcolor{bubbles} CL and RG &\cellcolor{bubbles} Seen-Seen                 &\cellcolor{bubbles} Seen-Unseen                          &\cellcolor{bubbles} $\surd$                &\cellcolor{bubbles} $\surd$   \\ \cline{2-7}
                                            &\cellcolor{bubbles} Reinforcement Continual Learning (1995)           &\cellcolor{bubbles} CL, CR and CNT    &\cellcolor{bubbles} Seen-Seen /Unseen                 &\cellcolor{bubbles} Seen-unseen          &\cellcolor{bubbles} $\surd$                &\cellcolor{bubbles} $\surd$             \\ \cline{2-7}
                                            &\cellcolor{bubbles} Continual Learning in Deep Neural Networks (2002) &\cellcolor{bubbles}  CL, CR and Rl &\cellcolor{bubbles} Seen-Seen / Unseen                &\cellcolor{bubbles} Seen-Unseen / Unseen                         &\cellcolor{bubbles} $\surd$            &\cellcolor{bubbles} $\surd$          \\ \cline{2-7}
                                            &\cellcolor{bubbles} Unsupervised Continual Learning (2014)            &\cellcolor{bubbles} CR and AS   &\cellcolor{bubbles} Unseen               &\cellcolor{bubbles} Seen-Unseen and Unseen        &\cellcolor{bubbles} $\surd$                &\cellcolor{bubbles} $\surd$  \\ \cline{2-7}
                                            &\cellcolor{bubbles} Semi-Supervised Continual Learning (2015)         &\cellcolor{bubbles} CR and CL &\cellcolor{bubbles} Seen-Seen and Unseen &\cellcolor{bubbles} Seen-Unseen          &\cellcolor{bubbles} $\surd$                &\cellcolor{bubbles} $\surd$             \\ \hline

\textbf{OWML}                        &\cellcolor{eggshell} Open-world Machine Learning (2015)                &\cellcolor{eggshell} CL, CR                &\cellcolor{eggshell} Seen-Seen/Unseen                 &\cellcolor{eggshell} Seen-Unseen and \textbf{Unseen-Unseen}/Unseen &\cellcolor{eggshell} $\surd$                &\cellcolor{eggshell} $\surd$ \\    \hline              
\end{tabular}
\textbf{Abbreviations:} ML: Machine learning, DL: Deep Learning, CML: Continual Machine Learning, OWL: Open-world Learning, CL: classification, RG: Regression, CR: Clustering, AS: Association, CNT: Control, RL: Representation learning
\end{table}

Traditional machine learning has five major tasks: classification, regression, association, clustering, and control (robotics), which are done with various kinds of Machine Learning (ML), which are shown in Table \ref{MlvsCML}.
Supervised machine learning proposed in the early 1988s uses seen-seen data for training and testing. In contrast, unsupervised Machine learning, which is also proposed in the 1980s, uses unseen data for training and testing. Semi-supervised machine learning uses seen-unseen data for both training and testing. Reinforcement learning, recommended for classification and control, perceives and understands its context, takes actions and acquires knowledge by experiments and oversights, uses the seen data for training and seen-unseen data for testing.
Deep learning is working on both classification and clustering uses seen data for both training and testing.
The task of all categories of continual machine learning and training and testing data are similar to similar traditional machine learning, except
traditional machine learning neither accumulates knowledge nor retains any previous knowledge in any future task.
Table \ref{MlvsCML}  given the broad classification of tradition and continuous machine learning based on task, required training and testing data, and knowledge accumulation and retention. Open-world machine learning, which uses seen data for training and seen, seen-unseen, and unseen data for testing, is the only method that has a rejection capability for unseen instances. 

\begin{figure}[htb]
    \centering
   \includegraphics[height = 2.0 in, width = 4.8 in]{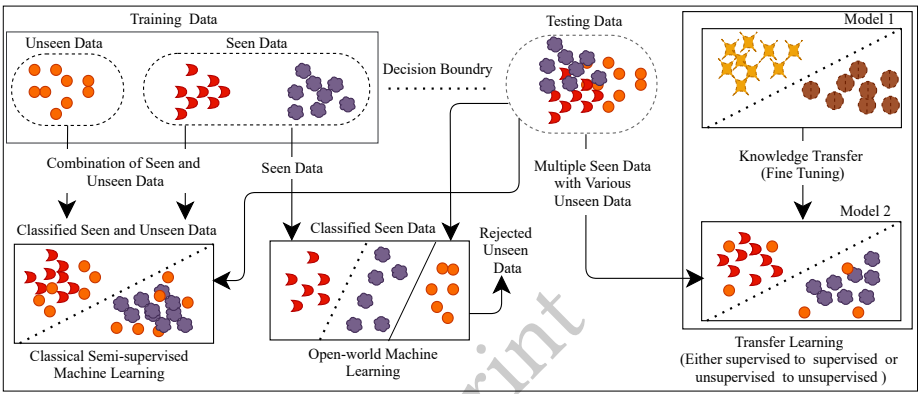}
    \caption{ Classical Semi-supervised
Machine Learning \textit{vs} Open-world Machine
Learning \textit{vs} Transfer Learning}
    \label{comparison}
\end{figure}

Sometimes it might appear that, open-world machine learning is associated with semi-supervised learning or transfer learning. However, these are different methods. Figure \ref{comparison} shows the comparison between semi-supervised machine learning, open-world machine learning, and Transfer Learning.  Semi-supervised machine learning involves small number of labelled data and possibly large number of unlabelled data. However, it still follows the closed world assumption that the unlabelled (unseen) data belongs to seen classes. It classifies instances according to classes available in training data. In contrast, open-world machine learning trained with seen data and classified seen data and rejected unseen data. Transfer learning uses knowledge transfer and fine-tuning to classify the new data (knowledge gained from one model can be used in different models to classify instances). Subsequently it works on new data and assume the closed world assumption during testing time.  For rest of the paper, we assume that seen classes means the classes which were appeared during training time, and unseen classes mean unseen-unseen classes.


Open-world machine learning problem can formally be defined as follows.

\begin{definition}
   \textit{ Let  D = \{$(x_1,\ y_1),(x_2,\ y_2), \dots, (x_i,\ y_i), \dots, (x_n,\ y_n )$\}, where $n$ is the total number of instances, is the labeled training data for $m$ seen classes. Here $x_i$ is the $i^{th}$ instance and $y_i \in $ \{$s_1$, \dots, $s_m$\}= $S$ is $x_i$'s class label.
   The objective of the classifier is to classify each test example $x$ to one of the $m$  seen classes or identify it as unseen class.}
\end{definition}

The  process of learning in open-world machine learning can be defined in following three steps.

\begin{itemize}
    \item Step 1: At specific time $t$, classification model builds by a learner that is multi-class classifier $M_t$ based on all previous classes $t$  of data with class labels $S^t = ( s_1,s_2, \ldots, s_t )$. $M_t$ is capable enough either classify seen classes $s_{i}  \in  S^{t}$  or reject them as  unseen classes and put them in a rejection set $R_e$. The $R_e$ may have instances of more then one new or unknown classes.

    \item Step 2: Now, the system can identify the hidden classes $c$ in $R_e$ and prepare training sets from this data to find unknown classes.

    \item Step 3: The model will learn from updated training dataset (previous data + new identified dataset). The model $M_t$ is update to a new model $M_{t+c}$.
  
\end{itemize}

There are many domains in which open-world machine learning  is beneficial, such as image processing, computer vision, and text processing. open-world machine learning  can be a bridge for self-motivated learning systems. open-world machine learning  is a relatively new field of machine learning; as we analyzed, most of the work in this area has only been done in this decade. We divided the review work into two parts: open-world machine learning  in Computer Vision and Image Processing (CV-IP), and the other  also known natural language processing (NLP); we have explored the literature work in these two domains.

\section{Review Methodology}

The methodical review summarized in this paper was done by succeeding conventional review processes that ease understanding of domains of open-world machine learning. The steps involved to write in this review article are the historical timeline,  convoying the survey, describing the outcomes, discussing investigations and challenges, the reasoning of conclusions, and future direction.

\subsection{Review Plan}

Conveying a methodical study includes collecting initial analysis about conclusions. Typical methods of such surveys incorporate confirmation and contradiction of preceding claims, classification and examination of analysis gaps/challenges, and future direction for exiting research. There is a fundamental advantage of conveying a methodical review and beneficial for authors as it covers the information of the domain with data. The following steps are taken to complete this survey.

\begin{itemize}
    \item Steps of  Review Plan
    \begin{enumerate}
        \item Recognize the requirement for a methodical survey
        \item Frame an investigation query.
         \item Find tasks and methods around that investigation query.
    \end{enumerate}
    \item Steps of Review and Result Reporting
    \begin{enumerate}
        \item Explore the initial investigations
         \item Study the initial investigations for significance and relevance of domains
         \item Selection of methodologies of the initial investigations
        \item Integrate and abstract the extracted studies from initial investigations
         \item Describe and report results as it is  with suitable datasets
        \item Conclude the methodologies and investigation 
        \item Conduct analytical and tabular comparisons
        \item Write the methodological survey
    \end{enumerate}
\end{itemize}

\subsubsection{Investigation  Queries}

We have formed the following generic queries to pursue the results from the readers’ perspective. These are the standard parameter and findings that are required to understand any domain of research. Further, we prepare the entire draft according to the review plan and investigation queries.

\begin{enumerate}
    \item  What is the importance of learning in the open-world?
    \item How has machine learning grown in the last decade?
    \item What are the classifications of open-world machine learning?, and 
    \item How does it differ from traditional Machine Learning (ML)?
    \item Which domains are correlated with OWML and how OWML can help to improve these domains?
    \item What is the current status of research in OWML?
    \item What are the tasks of OWML?
    \item What are the methods available in OWML to handling open-world tasks?
    \item How many datasets are available to investigate or perform OWML research for explicit domains.
   \item What are the associated areas of OWML?
   \item What are the challenges in the field of OWML to learn in open environments for various domains?
   \item What are the future directions of research in OWML?
\end{enumerate}

\subsubsection{Sources of information and Selection Criteria}

There is a need for a comprehensive aspect to the boundless coverage for an immeasurable and helpful article. We have collected a piece of pertinent information and data before getting started with a comprehensive article. We have explored many articles and select profoundly associated articles only to include them in the review. To collect this data, we have used prominent electronic sources, which are listed in Table \ref{esources}.

\begin{table}[htb]
\caption{E-source of Information}
\centering
\scriptsize
\label{esources}
\begin{tabular}{l|l|l}
\hline
E-Sources & Content Type & Total Article \\ \hline
https://www.acm.org & Journal and Conference & 43  \\
https://www.springer.com/in & Journal and Conference & 89  \\
https://www.ieee.org & Journal and Conference & 179 \\
https://www.elsevier.com/en-in & Journal and Conference & 94  \\
https://www.tandfonline.com/ & Journal and Conference & 17 \\
https://www.jmlr.org & Journal and Conference & 26 \\
https://www.aaai.org & Conference & 35  \\
https://www.kdd.org & Conference & 23 \\ \hline
\end{tabular}
\end{table}

\paragraph{Supplementary Sources: }

Other than the mainstream sources of information, we have used many repositories and other e-resources.  These sources are helping us to provide additional information, technical and scientific reports, and analytical data to understand the domain. Some of the sources are listed below. 

\begin{enumerate}
    \item https://mitpress.mit.edu (Books and Article)
    \item https://citeseerx.ist.psu.edu (Article)
    \item https://www.semanticscholar.org (Article and Technical Reports)
    \item https://www.morganclaypool.com (Book)
    \item https://www.kdd.org (Article)
    \item https://www.sciencedirect.com (Article and Technical Reports)
    \item https://www.connectedpapers.com/ (Article)
    \item https://scholar.google.co.in (Article and Technical Reports)
\end{enumerate}

\begin{figure}[htb]
\centering
  \includegraphics[height = 1.5in, width = 2.5in]{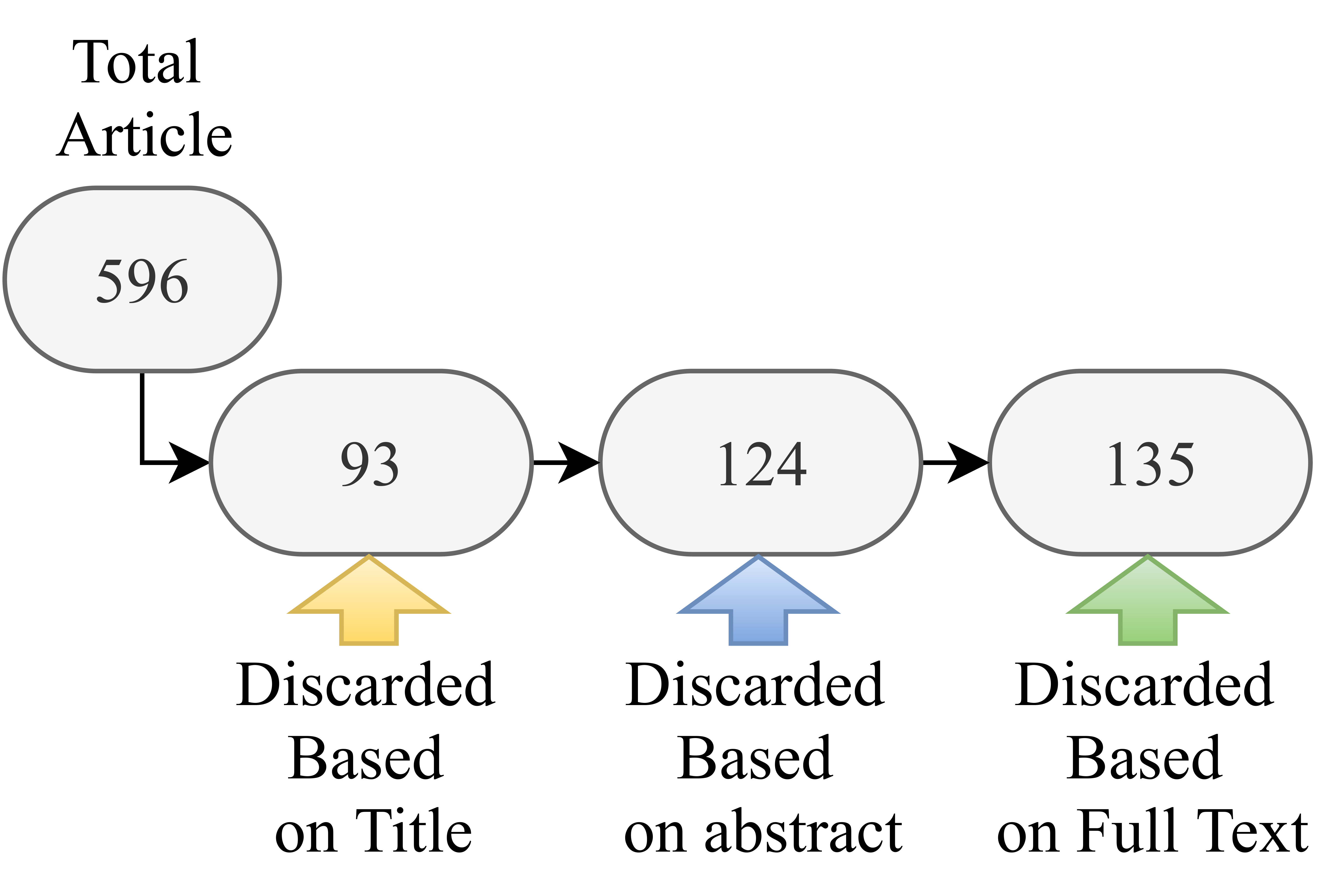}
   \caption{Article Discard Process}
    \label{SalectionCriteria}
\end{figure}

\paragraph{Article Search and Inclusion Criteria: } 
In approximately all searches carried the keyword ‘‘open-world’’ in its title or abstract, we keep it in our repository. The domain is relatively new, and most of the work has been done only in the last decade, so we have to access multiple sources to collect the information. We have detailed examined these articles and kept the relevant articles only, process shown in Figure \ref{SalectionCriteria}. Other than the keyword, we have used the most recent articles and technical reports to follow the rooted trail and find many relevant articles. To maintain the high authenticity of any claim, we used only notable and reputed sources articles in this review. 

\begin{figure}[htb]
\centering
  \includegraphics[height = 2.5in, width = 5.5in]{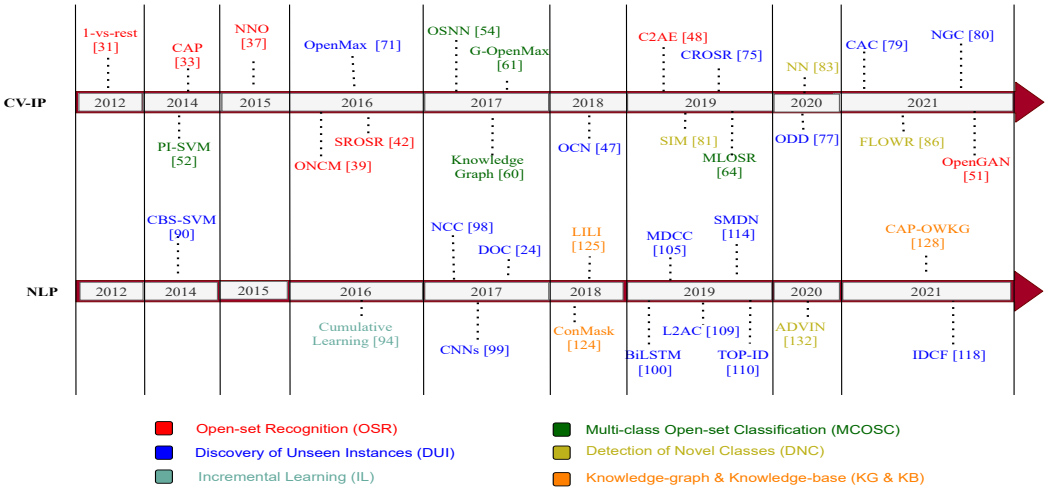}
   \caption{Timeline of Task Done in OWML for Both CV-IP and NLP}
    \label{timeline}
\end{figure}

After selecting articles by applying all the criteria on obtained articles from various sources, we have comprehensively studied the selected article on open-world machine learning.  Based on the study, articles are categorized in two significant domains of OWML, that is Computer Vision and Image Processing (CV-IP), and Natural Language Processing (NLP).
Further, the articles are classified based on task, some of them shown in Figure \ref{timeline}, and the rest are discussed in the following sections.

The timeline for open-world machine learning for both CV-IP and NLP shows in Figure \ref{timeline}. The classified task shown in the figure is categorized based on a domain with proposed years. The most of researches were done between 2016 to 2021. In the timeline, we have included only essential methodologies that have enhanced existing work or introduced novel methodology for specified tasks of open-world machine learning. We can observe that there is less research available in NLP for open-world machine learning than CV-IP. The timeline clearly shows that most of the research done in open-world machine learning is generally associated with discovering unseen instances. Some of the research focuses on the detection of novel classes for both CV-IP and NLP.

\subsection{Taxonomy of Open-world Machine Learning}

\begin{figure}[htb]
  \includegraphics[height = 2.5in, width = 5.5in]{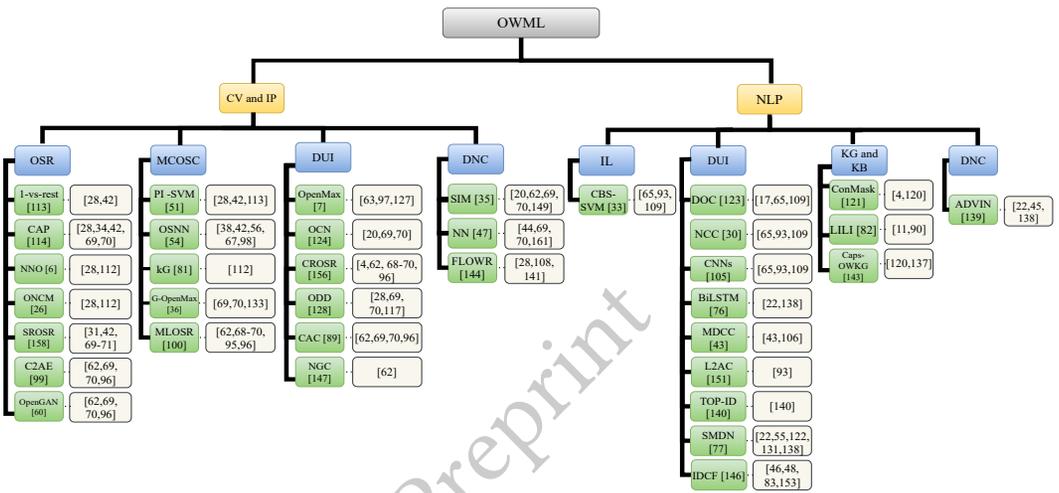}
   \caption{Taxonomy of Open-world Machine Learning}
    \label{Taxonomy}
\end{figure}

To ease the understanding of readers, we have graphically summarized (Figure \ref{Taxonomy}) the entire work done in open-world machine learning mentioned in this article. It involves cataloging of the domain, used or proposed methods, and dataset. There are two major fields where work has been done: computer vision \& image processing (CV-IP) and natural language processing. We have further categorized the work done in CV-IP based on tasks, such as Open-set Recognition (OSR), Multi-class Open-set Classification (MCOSC), Discovery of Unseen Instances (DUI), and Detection of Novel Classes (DNC). In computer vision and image processing, Numerous approaches have been used with various datasets to evaluate methods with different evaluation parameters.

 In computer vision and image processing to achieve various task 1-vs-rest, Nearest Non-Outlier (NNO), Extreme Value Theory (EVT), Conditioned Auto-encoder (C2AE), and Deep Neural Network (DNN) has been used, and these methods are evaluated with various datasets. The baseline benchmark algorithms are improved or extended to integrate the existing changes, such as Support Vector Machines (SVM) used to minimize open-space risk. Some other methods were also used for open-space risk minimization, such as NNO. The  Probability of Inclusion- Support Vector Machines (PI-SVM), Open-Set Nearest Neighbor (OSNN ) model, and EVT are used in many frameworks for multiset reorganization. The unseen class identification OpenMax, Generative OpenMax (G-OpenMax), Convolutional Neural Network (CNN), Classification-Reconstruction Learning for Open-Set Recognition (CROS), and DNN has been used. CNN and  Stream Classifier with Integral Similarity Metrics (SIM) have also been used to discover new classes.

We also discussed the work done in natural language processing in open-world; and further categorized it in Incremental Learning (IL), Discovery of Unseen Instances (DUI), Knowledge-graph \& Knowledge-base (KG \& KB), and Detection of Novel Classes (DNC).

 Several researchers have been used baseline algorithms such as Center-Based Similarity Support Vector Machines (CBS-SVM) in natural language processing to reduce the open-space risk and incrementally acquire knowledge. Several methodologies mentioned here are a significant part of the framework or are used standalone for unseen class discovery. The 1-vs-rest, CBS-SVM, Nearest Centroid Class (NCC), Long Short-term Memory Networks (LSTM), Learning to Accept Classes (L2CA), SoftMax, and Deep Novelty (SMDN), and  Automatic Discovery of Novel Intents (ADVIN) were used and evaluated with different datasets. The Lifelong Interactive Learning and Inference (LILI) model is used for new class detection, and OpenMax based models are used for text classification in the open-world.  Now we discuss the reviews of methodologies that were used for computer vision and image processing in detail (Section \ref{Section 4}). Next, we discuss methodologies for natural language processing in open-world machine learning  (Section \ref{Section 5}).

\section{Open-world Machine Learning  in Computer Vision and Image Processing} \label{Section 4}

In this segment, we discuss the literature works done in computer vision and image processing with open-world settings. The preliminary research is focused on open-set reorganization using various methods. The input images are unknown for the model, or we can say that input is novel and unseen; as the input images are not available in training data, the knowledge is incomplete for the model. The model needs to respond to these unseen (open) data. The task analysis of work done for the particular task is shown in Table \ref{taskwiseCVIP}.  Further, These tasks are discussed in Section \ref{osr}, \ref{mcosc}, \ref{cvipdui}, and \ref{cvipdnc}.   

\begin{table}[htb]
\caption{: Summarized Study of Task  Performed by Open-world Machine Learning in CV-IP}
\centering
\scriptsize
\label{taskwiseCVIP}
\begin{tabular}{c|c|c|c|c}
                                                                       \hline \textbf{Author}   & \textbf{OSR}          & \textbf{MCOSC}   &\textbf{ DUI}     & \textbf{DNC}     \\ \hline 
W.J. Scheirer et al.~\cite{scheirer2012toward}      & $\surd$        & -       & -       & -       \\ \hline
W.J. Scheirer et al.~\cite{scheirer2014probability} & $\surd$              & -       & -       & -       \\ \hline
L. P. Jain et al.~\cite{jain2014multi}              & -              & $\surd$ & -       & -       \\ \hline
A. Bendale and T. Boult~\cite{bendale2015towards}   & $\surd$              & -       & -       & -       \\ \hline
A. Bendale and T. Boult~\cite{bendale2016towards}   & -              & -       & $\surd$ &         \\ \hline
R. De Rosa et al.~\cite{de2016online}               & $\surd$        & -       & -       & -       \\ \hline
H. Zhang and V. M. Patel~\cite{zhang2016sparse}      & $\surd$        & -       & -       & -       \\ \hline
P. R. M. Junior et al.~\cite{junior2017nearest}      & -              & $\surd$ & -       & -       \\ \hline
S. Demyanov et al.~\cite{ge2017generative}          & -              & $\surd$       & - & -       \\ \hline
V. Lonij et al.~\cite{lonij2017open}                & -              & $\surd$ & -       & -       \\ \hline
L. Shu et al.~\cite{shu2018unseen}                  & -              & -       & $\surd$ & -       \\ \hline
P. Oza and V. M. Patel~\cite{oza2019c2ae}           & $\surd$        & -       & -       & -       \\ \hline
R. Yoshihashi et al.~\cite{yoshihashi2019classification} & -         & -       & $\surd$ & -       \\ \hline
P. Oza and V. M. Patel~\cite{oza2019deep}           & -              & $\surd$ & -       & -       \\ \hline
Y.  Gao et al.~\cite{gao2019sim}                    & -              & -       & -       & $\surd$ \\ \hline
M. Hassen and P.K. Chan~\cite{hassen2020learning}   & -              & -       & -       & $\surd$ \\ \hline
L.  Song et al.~\cite{song2020critical}             & -              & -       & $\surd$ & -       \\ \hline
 D. Miller et al.~\cite{miller2021class}            & -              & -       & $\surd$ & -       \\ \hline
J. Willes et al.~\cite{willes2021bayesian}          & -              & -       & -       & $\surd$       \\ \hline
 S. Kong and D. Ramanan~\cite{kong2021opengan}      & $\surd$        & -       & -       & -       \\ \hline
Z.-F. Wu et al.~\cite{wu2021ngc}                    & -              & -       & $\surd$ & -       \\ \hline
\end{tabular}
\end{table}

\subsection{Open-set Recognition (OSR)} \label{osr}

In a real-world environment, several external circumstances restrict the identification and distribution of tasks as inputs change frequently. It is generally challenging to accumulate training examples to employ all levels when training a  classifier. A more practical situation is open-set recognition occurs, wherever inadequate system information exists during the training, and unseen classes can be provided to a system during testing. In such a situation, expect the classifiers to correctly label the seen classes and effectively deal with unseen classes. Some research approaches this obstacle and recognizes open-sets.

In~\cite{scheirer2012toward}, the authors proposed an algorithm that can accept input with incomplete knowledge. The existing algorithm cannot handle open-sets. Thus they improve algorithms with the normalization of an algorithm to handle open-sets. They introduced 1-vs-set and performed experiments on Caltech-256~\cite{griffin2007caltech} and ImageNet~\cite{deng2009imagenet} sets. They perform experiments on labeled face data and compared it with their work. Also, perform experiments on different image domains and compare results with binary SVM linear kernel, binary 1-vs-set machine linear kernel, 1-class SVM linear kernel, and 1-class 1-vs-set linear kernel. Based on performance evolution, F-measure and accuracy clearly state that the binary 1-vs-set machine linear kernel performed better than the other algorithms.
Object and face recognition are considered in their work to verify the experimental results. Many researchers use multi-class classification to handle open-set problems, but multi-class approaches need labels for each input class. Therefore, the entire dataset required very laborious labeling and is still not an acceptable solution to handle open-sets. In this work, openness formalizes as:

\begin{equation}
   Openness=\sqrt{\frac{2* |T_{c}|}{|T_{s}|+T_{g}}}
\end{equation}

Where $T_{c}$= Training Class, $T_{s}$= Testing Class and $T_{g}$= Target Class. It yields openness in percentage between 0 to 100. Where 0= complete close class and 100  denote maximum openness, they conduct experiments on SVM with half-space and classify data not available in the training set. The SVM found separate classes as negative and positive, and the negative is only for known objects. The Rest of the unknown objects are left as large unclassified open-space, which could be a part of a positive set. Then they felt that this could be remedied by reducing the open-space. Instead of generalization and specialization to minimize the errors in training function, they introduce “set”. Set is known class training data of 1-vs-set. It is used for open-space risk models and error minimization.

In~\cite{scheirer2014probability}, the authors proposed a 1-vs-rest machine. The 1-vs-Set algorithm handles the hazard of the unknown classes by dealing with two plane optimization, likely to result as a linear classifier. They extended open-risk classification to include non-linear classification in multi-class settings. They suggested a new model Compact Abating Probability (CAP) based on Weibull-calibrated SVM (W-SVM); it decreases the probability value of member class when points move towards open-space from known data. The CAP evaluated on publically available benchmark datasets Letter~\cite{frey1991letter}, MNIST~\cite{lecun1998mnist,lecun1998gradient}, Caltech-256~\cite{griffin2007caltech}, and ImageNet~\cite{deng2009imagenet}. The results show that the CAP can reduce the open-space risk for known data.  In~\cite{scheirer2012toward}, it is established that by optimizing two planes, the 1-vs-set machine can manage the risk and produce a linear classifier. 1-vs-rest reduces the open-space risk by interchanging half-space but still, the open-space risk is infinite. For certain known classes, it is quite easy to find known but when handling multisets. Here three basic categories of the class are denoted, known classes (the class with abstractedly labeled, that is a positive example), known unknown classes (negative examples), and unknown classes (unseen classes) Figure. The algorithm is specially designed for unknown classes, and this algorithm deduced open-space risk from infinite to finite.

1-vs-set machine assigns class labels to examples through the testing. It used a probability decision score for multi-class. It classifies examples by multiple classifiers based on the highest probability or probability that goes beyond the threshold. Examples that are below the threshold are rejected as unknown. This work has formalized a compact abating probability to address open-set regeneration by introducing a new algorithm, W-SVM, which integrates compact abating probability model and probability estimation theory. The experimental results clearly state that the f-measure (openness) of the W-SVM is relatively high for both open-set binary object regeneration and multi-class open-set recognition. The W-SVM also performed decently on the OMNIST dataset for multi-class open-set recognition.

In~\cite{bendale2015towards}, authors address the issuers associated with open-world recognition, such as open-space risk and practical tasks.  They proposed a protocol to evaluate open-world recognition. They proposed Nearest Non-Outlier (NNO) algorithm to manage open-space risk, model efficiency, and adding object categories incrementally while detecting outliers. The proposed NNO algorithm experiment on more than 1.2 million images of ImageNet~\cite{deng2009imagenet} dataset to validate the model. NNO is an extension of the Nearest Class Mean (NCM)~\cite{ristin2014incremental} algorithm. They set an open-world evaluation protocol in which uses seen classes in training. However, both seen and unseen classes are used during the testing and continually add new class categories when encountering unseen classes. The training phase is further divided into two phases, the first is metric learning, and the second is the incremental learning phase.    

NNO is a combination of two open-space risks for model combination and open-space risk for threshold space. NCM~\cite{ristin2014incremental} assumes that all classes are known that are not suitable for open-world recognition; hence NNO extends its feature and considers a measurable recognition function. It is given the probability of an object being in any class. It is always greater than zero. When a measurable function is zero, and all classes reject, the NNO rejects the object as unknown. After collecting all unknown novel objects, NNO considered these objects as new classes. NNO can identify unknown objects and include these unseen classes as new classes. Hence NNO is an open-world algorithm. The experiment has been done with two different sets of categories that are 50 and 200 and applied SVM, 1-vs-set, NCM, NNO. The experiment evaluations state that NNO outperformed the rest of the algorithms and generate significantly better outcomes when testing with unknown categories of classes for both 50 and 200 sets of categories.

In~\cite{de2016online}, an author extends the work of open-world recognition~\cite{bendale2015towards}. They argue that to capture dynamic word recognition, needed incremental learning of underlying matrices, confidence threshold for unseen classes, and description space of class. They conduct experiments in three phases, first large-scale increment learning, second is open-world recognition, and third is an online prediction of streamed images. The ImageNet~\cite{deng2009imagenet} and ILSVRC’10~\cite{russakovsky2015imagenet} data set used in experiments, which consist of 1.2 million, 50K, and 150K images for training, validation, and testing, respectively. The large-scale increment learning method created relevant matrices and learned parameters on the initial set of 20 classes, then classes are added incrementally in the set of 10 classes. To recognize classes in an open-world environment method, learn parameters and metrics on an initial 50 classes, and images of 50 classes are added after each iteration to evaluate the performance on a test set that comprises both known and unknown classes. To predict online images, researchers used current methods Nearest Class Mean (NCM) classifier and Nearest Ball Classifier (NBC) and updated these accuracy. They predict images using ONCM and ONBC and compare results with existing NCM and NBC. Initially predicted the labels for samples with the current model, then update online accuracy based on predicted labels and genuine labels. After updating the online accuracy, updated the existing methods using accurate labels and sets. They also conduct the same experiment on Places-2 dataset~\cite{zhou2017places}. The results clearly state that the ONNO, ONCM, and ONBC are performed better than existing algorithms.

In~\cite{zhang2016sparse}, the authors proposed a framework that works on sparse representation-based classification (SRC). SRC used class reconstruction error for classification. The most useful information about open-sets is available in the tail of the similar and non-similar parts of the class. To tail distribution reconstruction error SRC uses statistical Extreme Value Theory (EVT)~\cite{de2006extreme}. To evaluate the result they used benchmark datasets, such as extended Yale B~\cite{lee2005acquiring}, MNIST~\cite{lecun1998mnist,lecun1998gradient}, UIUC Attribute dataset~\cite{farhadi2009describing} and Caltech-256 Dataset~\cite{griffin2007caltech}. Evaluation results show that the simple, sparse representation classification does not effort up to the mark for open-sets. Hence introduce different training modules to the trained system to handle open-sets. In training, acquired a random sample for each class then partition the class into two sets, cross-train and cross-test. These partitions are used for training and testing. Cross-train contains 80, and cross-test contains 20 percent of training samples. To evaluate the result of sparse representation-based open-set recognition (SROSR), Compare the result with existing methods that used sparsity-based rejection such as w-SVM, Sparsity Concentration Index (SCI)~\cite{wright2008robust}, ratio, and Naïve. The result clearly states that the SROSR is providing more accuracy and better F-Measure than existing methods.

The encoder-based model~\cite{shu2018unseen} to detect unseen classes are further extended and use encoder and decoder for classification and open-set detection. In~\cite{oza2019c2ae}, the authors proposed a C2AE method that used class conditioned auto-encoder to recognize open-set with novel training and testing methodologies. The proposed model worked in two parts, close-set classification and open-set identification. Encoder learned the first task for close-set classification, and decoder learned the next task for open class identifications. Training has been done using a close-set model. The close-set model consists of known classes. It trained encoder and classifier and conventionally calculated classification loss; after training encoder for close-set, it trained open-set identification module which consists of an auto-encoder network with weight and decoder for the reconstruction of the image according to label condition vectors. Testing of the model has been done for an open-set using a k-inference algorithm. To evaluate the performance of the model for open-set compare the model with W-SVM~\cite{scheirer2012toward}, SROR~\cite{zhang2016sparse} and DOC~\cite{shu2017doc}, on MNIST~\cite{lecun1998mnist,lecun1998gradient}, SVHN~\cite{netzer2011reading} and CIFAR10~\cite{krizhevsky2009learning} benchmark datasets.

In~\cite{kong2021opengan}, the authors proposed an OpenGAN to recognize the open-sets by open data generation. The OpenGAN consists of the GAN-discriminator to classify testing examples. It is a binary classifier trained for both open-set and closed-set data.  There are many other techniques available for close-set classification, but each technique has limitations, but OpenGAN overcomes them by integrating them with various technical insights. In the first step, OpenGAN picked GAN Discriminator on a few actual outlier data already used in the existing research. The second step synthesizes "fake" data and adds it to the complete open training examples. The proposed method methods are evaluated with benchmark datasets (CIFAR, SVHN, and MNIST) and show promising outcomes for open-set recognition vi open data generation.


\subsection{Multi-class Open-set Classification (MCOSC) } \label{mcosc}

Multi-class classification in the open-world is a very challenging task. If the unknown classes remain unaddressed, the classifier either misclassifies or classifies false, known classes, and there is also a possibility to classify false unknown classes. Misclassification and false classification can be avoided if multi-class classifiers can identify unknown classes appropriately.

In~\cite{jain2014multi}, authors articulate the problem as one of sculpting positive training data at the decision boundary and invoke the arithmetical theory. The new algorithm termed PI-SVM recalls higher accuracy than existing. It is used for assessing the non-normalized posterior likelihood of class insertion. They convert MNIST~\cite{lecun1998mnist,lecun1998gradient} data set from closed set to open-set recognition task and experiment with different sets of training and testing data. Now the following steps have been followed: (i) It used the standard supervised learning algorithm for MNIST~\cite{lecun1998mnist,lecun1998gradient} classification with a 1-vs-rest SVM on Platt~\cite{platt1999probabilistic} probability estimation in which classes are seen during the training, (ii) Used only six classes from MNIST, (iii) Used all ten classes from MNIST and four unseen classes during the training,  and (iv) Change the testing regime to cross-class validation. In this scenario, similar classes are held out during the training (it is just a shuffling in (ii)) but comprise in a testing (it is a shuffling in (iii)).
They have performed two distinct open-set scenarios with the cross-class validation: object detection by specific classifiers and multi-class open-set recognition followed by detection of a problem and then compares it with PI-SVM.  To evaluate performance measures and binary decision elements for an open-set object decision used a universe of 88 classes~\cite{scheirer2012toward}. 
To train the model, used images from Caltech-256 and for testing took the images from both Caltech-256 and ImageNet~\cite{deng2009imagenet}. The entire evaluation was done over the five fold cross dataset. The result clearly states that PI-SVM improves the F-Measure by 12 to 22 percent compared to existing methods.

In~\cite{junior2017nearest}, the authors proposed Open-set Nearest Neighbor (OSNN) to address the issues in multi-class classifiers. It is an extension of the Nearest Neighbor (NN)~\cite{keller1985fuzzy} for open-set. OSNN used a similarity ratio instead of a similarity score and applied a threshold to find similarities between classes. They also designed a specific experimental protocol to evaluate open-set methods. Earlier proposed algorithms and frameworks are displayed as a virtuous outcome in the experiment, but in real-world applications, these algorithms and systems are straggled to perform with open-set. Hence, to overcome these issues, they also proposed a system that measures adaptation in an existing open-set classification system, that is, Normalized Accuracy (NA) and Open-set F-measures (OSFM), and evaluates the classifier's performance for both seen and unseen classes. The evaluation done on 15-Scenes~\cite{lazebnik2006beyond}, Auslan~\cite{kadous2002temporal}, Caltech-256~\cite{griffin2007caltech},  ALOI~\cite{geusebroek2005amsterdam} and   Ukbench~\cite{nister2006scalable} datasets.

Visual recognition systems play an essential role in identifying both seen and unseen classes of images. In~\cite{lonij2017open}, authors proposed a knowledge graph-based approach to identify unknown visuals and recognize visuals in the open-world. Three basic methods are used to predict classes, first Standard classification settings can predict only classes that are available in training, and images that are not available in training cannot be accurately predicted by standard classification. The second, zero-shot setting can predict images that are not available in training, but some partial information is available for novel classes. The third open-world setting can predict images that are neither available in training nor any partial information about its classes. The proposed method used the knowledge graph embedding model and image embedding model. The knowledge graph model uses properties, and the image embedding model uses images for training (ILSVRC-2012~\cite{russakovsky2015imagenet}). Image embedding models used properties of the knowledge graph to predict open-world images.

In~\cite{ge2017generative} authors proposed  Generative OpenMax (G-openMax), which calculated the decision score of unseen classes instead of seen classes. It is an extension of OpenMax, which consists of a GANs~\cite{goodfellow2014generative} network. The proposed method used visualization for both seen and unseen classes; it also used probability estimation to GANs, and previous seen class dissemination to produce reasonable and domain-adapted synthetics unseen samples. The evaluation has been done on both small and large scales datasets. A minimum of 10, and a maximum of 95 classes are utilized for the openness problem on two (HASYv2 dataset~\cite{thoma2017hasyv2} and MNIST~\cite{lecun1998mnist,lecun1998gradient}) handwritten datasets.

In~\cite{oza2019deep}, authors’ proposed Multi-task Learning-based Open-Set Recognition (MLOSR). It is based on a neural network for multitasking in open-set visual recognition. The proposed method is a combination of a classification network, decoder network, and feature extractor network. It utilized a decoder network to reject an open-set, and the decoder network reconstructs the error. It also uses EVT~\cite{de2006extreme} for model tail error reconstruction from seen classes. EVT improves the overall performance of the model. The feature extractor network took input and generated the latent. The classifier uses this latent decoder to predict the class labels and reconstruction of input images. The entire network has trained for both reconstructions of input images and classification. EVT modeled the trail of the reconstruction of the error distribution. The probability of reconstruction error by EVT and classification score used for open-set recognition testing. Experiment done on COIL-100~\cite{nene1996columbia}, MNIST~\cite{lecun1998mnist,lecun1998gradient}, SVHN~\cite{netzer2011reading}, CIFAR10~\cite{krizhevsky2009learning} and Tiny-ImageNet~\cite{le2015tiny} datasets, MLOSR are tested with benchmark network VGG and dance Net with SoftMax~\cite{goodfellow2016deep}, OpenMax~\cite{scheirer2011meta} and combination of ladder net, DHRNet with SoftMax, OpenMax, and CROSR, MLOSR performed better than an existing network to recognize open-set.
Existing work focused on the prediction of unknown classes. The identification of newer classes from unknown classes is difficult without finding instances of unknown classes. 

\subsection{Discovery of Unseen Instances (DUI)} \label{cvipdui}

Generally, systems choose images that might not be useful or significantly meaningless. In traditional classification methods, the system has to classify the testing object in some of the classes. In comparison, an ideal system must reject the unseen classes that are meaningless and irrelevant. Some of the work presented shows how “fooling”~\cite{nguyen2015deep} and “rubbish”~\cite{goodfellow2014explaining} images appear in relevant classes as their confidence is high, whereas these are far from the class in which they appeared. Traditional deep networks have used fully connected feeds to the SoftMax layer~\cite{goodfellow2016deep} as output. SoftMax produces probability for the known labeled classes.

In~\cite{bendale2016towards}, the authors addressed this issue by introducing a methodology that can reject the unseen classes while testing. It is an adapted deep network for open-set identification. This methodology introduced OpenMax~\cite{scheirer2011meta} that can evaluate the likelihood of an input being for an unseen class. OpenMax rejects unrelated images, reduces the error rate, and manages open-space risk. OpenMax estimates class by measuring a distance between the model vector aimed at the limited upper classes and the activation vector for an input. OpenMax provides the likelihood of unknown classes. Here OpenMax has an extended version of SoftMax that includes probability for unknown classes. This method used meta-recognition in deep networks and found scores to estimate how far testing an object to a known class. To estimate the score activation layer has been used in deep networks. Meta-recognition and OpenMax can differentiate seen and unseen classes and avoid foolish images to classify in known classes. The proposed model was evaluated on ImageNet, a subset of the ILSVRC-2012 dataset since ILSVRC-2012 test labels are unavailable for use,  experiments stated on validation set performance~\cite{nguyen2015deep,krizhevsky2012imagenet,simonyan2014very}.

To extend neural network-based unseen class discovery and add the capability of rejection combination of networks has been used. In~\cite{shu2018unseen}, authors proposed a framework to identify seen classes and reject unseen classes, seen classes that are available in training, and unseen classes which are available at the time of testing. It is not possible without having previous knowledge. The objective is to discover unseen classes for any given task and make a cluster by rejected examples. Open-world machine learning is quite different from knowledge transfer. In the knowledge transfer mechanism, the system sends information between supervised to supervised and unsupervised to unsupervised systems. In this work, knowledge is shared from supervised to unsupervised. To achieve the objective, they develop a model that consists Combination of two networks an Open Classification Network (OCN) and a Pairwise Classification Network (PCN). Both networks will share the same omponents for learning. OCN is Build function F(x) that can classify each seen and unseen class in S where PCN Build $g(x_p,\ x_q)$, a binary classification model. PCN will identify two test examples seen, unseen, from the same class or different classes, and Hierarchical clustering used to discover hidden classes in all rejected examples. To test compare with existing work, the proposed model, evaluated on MNIST~\cite{lecun1998mnist,lecun1998gradient}, and EMNIST~\cite{cohen2017emnist} dataset. 

All the methods discussed above are trained in a supervised manner and designed to classify known classes that are available at the time of training. Therefore it is tough to determine unseen or unknown classes using these methods. It also upshots the accuracy of the classification of known classes.

In~\cite{yoshihashi2019classification}, authors proposed Classification-reconstruction Learning for Open-set Recognition (CROSR) for robust unknown classes deprived of distressing the classification accuracy of known classes. CROSR trained networks for categorization and restoration of input data. While learning to distinguish unseen from seen and classes of seen, this technique helps to improve the implicit interpretation. 
To provide durable unseen recognition despite compromising the efficiency of seen-class classification, CROSR method uses implicit structures for reconstruction.  CROSR is based on OpenMax formulation. It reconstructs the input data to detect uuseen classes. They use exclusionary learning algorithms in seen classes to build their classifiers. An open-set classification system based on DHRNets, CROSR combines seen classification with unseen detection.
This technique outperforms existing deep open-set classifier algorithms DOC~\cite{shu2017doc}, SoftMax~\cite{goodfellow2016deep} and OpenMax~\cite{scheirer2011meta}, for most permutations of seen data and anomalies, according to the trials conducted on five typical picture and text datasets MNIST~\cite{lecun1998mnist,lecun1998gradient}, CIFAR-10~\cite{krizhevsky2009learning}, SVHN~\cite{netzer2011reading}, tiny-ImageNet~\cite{le2015tiny} and DBpedia~\cite{auer2007dbpedia}. 

The current research scenario focuses on finding new classes in rejected data that are unseen or unknown. It will make the system more realistic and capable of working as a human being in a dynamic environment. In~\cite{song2020critical}, authors focused on the impact of out-of-distribution detectors and evaluated the performance of detectors. They took six Out-of-distribution Detectors (ODD) , which are published in the best conferences in the world. They also tested detectors for corrupt images that’s effect is unpredictable on the outcome; it may improve or decrease the performance. The out-of-distribution detectors ODIN, Network Agnostophobia, Mahalanobis Detector, Auto-encoder Detector, Deep-SVDD, and Outlier Exposure, are evaluated with MNIST~\cite{lecun1998mnist,lecun1998gradient}, VOC12, ImageNet, Internet Photos~\cite{sehwag2019analyzing}, Gaussian Noise and Uniform Noise with WRN-28-10 model using a different combination of in-distribution. The performance evolution states the adversarial training can improve the end-to-end strength. Adversarial training decreases discriminative influence and leads to poorer detection performance on benign out-of-distribution data.

In~\cite{miller2021class}, the authors proposed a simple Deep Neural Network (DNN) based framework for open-set classification. DNN contains Open-set Layer (OS-Layer) and Closed-set Layer (CS-Layer). It splits the data of intraclass. DNN splits data into subsets and produces an atypical sample. Atypical samples are used to model then abnormal data and normal samples are used for training. Intraclass info splitting exploits the inter-class information. The closed set regularization deep neural network apprehends an extraordinary close-set precision, and it is competent to discard unseen classes. The experiment performance evaluation is done on MNIST~\cite{lecun1998mnist,lecun1998gradient}, SVHN~\cite{netzer2011reading} and CIFAR10~\cite{krizhevsky2009learning} and compares results with WSVM, OCSVM, GAN, CF, and AE-ics.

In~\cite{wu2021ngc}, The authors present NGC, a novel graph-based noisy tag learning framework, which rectifies in-distribution noisy tags and filters out-of-distribution examples by leveraging the confidence of model predictions and geometric characteristics of the data, when it comes to testing. NGC can identify and discard out-of-distribution samples without any additional training. NGC is evaluated on CIFAR-10 and CIFAR-100  publically available benchmark datasets  associated with real-world tasks. The experimental evaluation of NGC shows that shows improvement over the existing methods.

\subsection{Detection of Novel Classes (DNC)} \label{cvipdnc}

The key challenge is finding instances of newly presented data known in nature for the system. Most researches are focused on data with a low dimension dependent on coherence data and its property; therefore, detecting instances for newly known classes is hard to detect.

In~\cite{gao2019sim}, the authors proposed a solution to this problem. The proposed framework SIM is a semi-supervised stream classifier that performs classification and detects novel classes on high-dimensional data streams. It uses latent features space for classification, and an open-world classifier implements metric learning, stream classification, and detects novel classes in unseen data. The performance evaluation was done on both image and text data. To test image model they calculate  novel misclassified
instance ($M_{new}$) and  existing instances misclassified as a
novel ($F_{new}$) apart from slandered performance majors FASHION-MNIST~\cite{xiao2017Fashion}, MNIST, EMNIST, and CIFAR-10 and for real-time text data, articles from the New York Times and Guardian have been used with ten classes of other news.

open-world machine learning  has also extended its significance in security as we have new kinds of malware in every period. To recognize that type of unseen class of malware, we need a system to detect undefined classes. In~\cite{hassen2020learning}, authors proposed a method that can detect new unseen classes of malware. In this exemplification, samples from a similar class are closed to each other while those from different classes are further apart, leading to more significant space between known classes for unknown class samples to occupy. The proposed algorithm uses three datasets to evaluate the results, MNIST~\cite{lecun1998mnist,lecun1998gradient}, MS challenge~\cite{guo2016ms}, and Android genom~\cite{zhou2012dissecting}.

 In~\cite{willes2021bayesian},the authors proposed open-world classification techniques that use embedding-based few-shot learning algorithms. It comprises small context and big context few-shot open-world recognition formalization where decision-making machines must classify existing classes. Few-shot learning for open-world recognition combines Bayesian non-parametric class priors with an embedding based pre-training method.  It also discovers unknown classes and then quickly adapts and generalizes classes with the limited labeled data. It adapts benchmarks approaches such as few-shot training, open-set classification, and open-world identification to this environment.  The authors present a Bayesian few-shot learning technique based on Gaussian embedding. The proposed system can efficiently integrate new classes for both few-shot open-world recognition situations and Bayesian non-parametric classes. The evaluation results show that the proposed approach improves on a range of current methodologies by 12 percent in terms of H-measure. They evaluate the proposed model on Mini ImageNet~\cite{vinyals2016matching} and TieredImageNet~\cite{ren2018meta} few-shot learning datasets(Subset of ImageNet ILSVRC-12~\cite{deng2009imagenet}).

Table 4 shows a summarized illustration of literature on open-world machine learning in computer vision and image processing.  It shows the used or recommended methodology, datasets employed for evaluation, and proposed results by the authors.

\begin{scriptsize}
\begin{table}[htb]
\caption{Summarized Study of Open-world Machine Learning in CV-IP}
\scalebox{0.9}{
\begin{tabular}{p{2.4cm}|p{2.0cm}|p{2.8cm}|p{6.0cm}}

\hline 
\textbf{Author(s)}                                       & \textbf{Proposed /Used  Methodology}      & \textbf{ Dataset}                  & \textbf{ Reported Results}             \\ \hline
W.J. Scheirer et al.~\cite{scheirer2012toward}           & 1-vs-rest                        & Caltech-256 and ImageNet & F1-score 80\%, Accuracy 98\% \\ \hline
W.J. Scheirer et al.~\cite{scheirer2014probability}      & Compact
Abating Probability (CAP)                                &  Letter, MNIST, Caltech-256, and ImageNet                    & F-measure 95 to 98\%  for 0 to 14\% Openness                              \\ \hline
L. P. Jain et al.~\cite{jain2014multi}                   & $P_{I}$-SVM                       &  Letter, MNIST, Caltech-256, and ImageNet                        & F-measure 88 to 98\% for 0 to 14\% Openness                             \\ \hline
A. Bendale and T. Boult~\cite{bendale2015towards}        & Nearest Non-Outlier (NNO)                       & ImageNet and ILSVRC'10                         &                              \\ \hline
A. Bendale and T. Boult~\cite{bendale2016towards}        & OpenMax                       & ImageNet (ILSVRC'10)                         & F-measure 0.59\% for Threshold values 0.20 to 0.45                              \\ \hline
R. De Rosa et al.~\cite{de2016online}                    & ONCM, ONNO, and ONBC                       &  ImageNet (ILSVRC'10)                         & Top-1 Accuracy 43\% for known Train Classes \newline Top-1 Accuracy 49\% for Unknown Train Classes (50 Known Classes)                            \\ \hline
H. Zhang and V. M. Patel~\cite{zhang2016sparse}          & Sparse Representation-based OPen-Set Recognition (SROSR)                       &  MNIST, Extended Yale B,UIUC attribute, and Caltech-256                         &   F1-measure 93 to 98\% for 0 to 14\% Openness \newline Accuracy 92 to 99\% for for 0 to 14\% Openness                              \\ \hline
P. R. M. Junior et al.~\cite{junior2017nearest}          &  Open-Set Nearest-Neighbor (OSNN)                       &  15-Scenes, Letter, Auslan, Caltech-256, ALOI, and Ukbench                        &   Normalized Accuracy 90\%(Max. with Auslan) \newline Micro open-set F-measure 80\% (Max. with Letter) \newline Closed Accuracy 90\% (Max. with ALOI)                           \\ \hline
S. Demyanov et al.~\cite{ge2017generative}               &  Generative OpenMax (G-OpenMax)                       & MNIST and HASYv2                         & F-measure 80 to 99\% for 0 to 13\% openness \newline Accuracy 58\% (Maximum with MNIST)                             \\ \hline
V. Lonij et al.~\cite{lonij2017open}                     & knowledge-graph                       & ILSVRC-2012                         & Fracrion of Image 85\% (With atleast 1 correct triple)   \newline Mean Rank 14\%, and \newline average number of true triples 19\%                          \\ \hline
L. Shu et al.~\cite{shu2018unseen}                       &  Open  Classification Network (OCN)
 CNN and 1-vs-rest                       & MNIST and EMNIST                         & Mirco F1-score 91\% (Max with MNIST) \newline Accuracy 81\% (Max with EMNIST)                             \\ \hline
P. Oza and V. M. Patel~\cite{oza2019c2ae}                & Class Conditioned Auto-Encoder (C2AE)                       & MNIST, SVHN, CIFAR10, CIFAR+10, CIFAR+50, and TinyImageNet                         & F-measure 82 to 94\% for 0 to 100\% openness.                          \\ \hline
R. Yoshihashi et al.~\cite{yoshihashi2019classification} & Classification-Reconstruction learning for Open-Set
Recognition (CROSR)                        & MNIST, CIFAR-10, SVHN, TinyImageNet, and DBpedia                         & F-Measure 41 to 79\% for the threshold value o.1 to 0.9 ( Maximum With MNIST) Micro F1-score 82.7\% (Maximum with CIFAR-10 )                             \\ \hline
P. Oza and V. M. Patel~\cite{oza2019deep}                & Multi-task Learning Based Open-Set Recognition
(MLOSR)                      & MNIST, SVHN, CIFAR10, CIFAR+10, CIFAR+50,  COIL-100, and TinyImageNet
                         & F-measure 82 to 90\% for 0 to 49\%                             \\ \hline
Y.  Gao et al.~\cite{gao2019sim}                         & Stream Classifier with Integral Similarity Metrics (SIM)                       & Image Datasets: Fashion MNIST, MNIST, EMNIST CIFAR-10              \newline Text Dataset:  NEW YORK TIMES, GUARDIAN                         & Image Dataset: Accuracy = 96.94\% Label Ratio = 100\% Effectiveness = 96.94\%  Mnew = 61.3\% Fnew = 47.1\% Text Dataset: Accuracy = 57.95\%  Label Ratio =96.0\%  Effectiveness = 57.95\%  $M_{new}$ = 62.14\%  $F_{new}$ =59.0\%                                       \\ \hline
M. Hassen and P.K. Chan~\cite{hassen2020learning}        & Neural-network                       & MNIST, MS Challenge, and Android Genom                         & AUC 95.88\% for 100\%FPR and 8.30\% for 10\% FPR (Maximum with MNIST)                             \\ \hline
D. Miller et al.~\cite{miller2021class}                  &  Class
Anchor Clustering (CAC)                       & MNIST, SVHN, CIFAR10, CIFAR+10/+50, and TinyImageNet                         & Area Under the ROC Curve (AUROC) 99.1\% (Maximum with MNIST)                               \\ \hline
J. Willes et al.~\cite{willes2021bayesian}               &  few-shot learning for open-world
recognition (FLOWR).                      & Mini ImageNet and TieredImageNet (Both are subset of ILSVRC-12)                         & Accuracy 51.64\% , Support-accuracy 57.76\% and Incremental-Accuracy 39.39\% (Maximum with Mini ImageNet) H-Measure 19.06\% (Maximum with TieredImageNet)                            \\ \hline
S. Kong and D. Ramanan~\cite{kong2021opengan}            & Open Generative adversarial networks (OpenGAN)                      & CIFAR, SVHN, MNIST, and Cityscapes                         & AUC 98.0\% (Maximum with CIFAR)   and F1-score 58.7\% (Maximum with Cityscapes)                          \\ \hline
Z.-F. Wu et al.~\cite{wu2021ngc}                         & Noisy Graph Cleaning (NGC)                        & CIFAR-100, TinyImageNet, and Places-365                         & Accuracy 94.18\% (Maximum with Places-365) AUROC 94.31\%  (Maximum with CIFAR)                           \\ \hline
\end{tabular}}
\end{table}
\end{scriptsize}

\subsection{Available Software Packages and Implementations}
In this section, we provided a link for some for software packages which contains the various implementation of various model of open-world machine learning  in computer vision and image processing (Table \ref{Softpack_CVIP}). These are the models commonly used in various frameworks of open-world machine learning .

Available software packages can be used to improve further learning in the open-world for computer vision and image processing. The 1-vs-rest is helping to improve the rejection of unknown classes. The Nearest Non-Outlier (NNO) can normalize the open-space risk and open-set reorganization. Conditioned Auto-encoder (C2AE) is an encoder and decoder method for open-set reorganization,  Multi-stage Deep Classifier Cascades (MDCC) for finding new classes.  The  Probability of Inclusion- Support Vector Machines (PI-SVM) and  Weibull-calibrated Support Vector Machines (W-SVM) can be used for multi-class classification in open-world machine learning .

\begin{table}[htb]
\centering
\caption{Available Software Packages and Implementations}
\label{Softpack_CVIP}
\scriptsize
\begin{tabular}{p{2.9cm}| p{2.2cm} |p{7.0cm}}
\hline
\textbf{Author} & \textbf{Model} & \textbf{Link}                                                 \\ \hline 
{W. J. Scheirer et al.~\cite{scheirer2012toward}}         & 1-vs-Set       & https://github.com/Vastlab/liblinear.git                      \\ \hline
{A. Bendale and T. Boult~\cite{bendale2015towards}}        & NNO            & http://vast.uccs.edu/OpenWorld                                \\ \hline
{P. Oza and V. M. Patel~\cite{oza2019c2ae}}        & C2AE           & https://github.com/dhruvramani/C2AE-Multilabel-Classification \\ \hline
{R. Yoshihashi et al.~\cite{yoshihashi2019classification}}        & CROSR          & https://nae-lab.org/$\sim$rei/research/crosr/                 \\ \hline
{C.-C. Chang et al.~\cite{chang2011libsvm}}        & W-SVM, PI-SVM  & https://github.com/ljain2/libsvm-openset.                     \\ \hline
\end{tabular}
\end{table}

\subsection{Discussion}
Many algorithms and frameworks are given significant outcomes for images in real-world settings. However still, there is a need for a generic framework to deal with real-time inputs in a dynamic environment. Ideal outcomes can be achieved if models can adopt generalization or specialization and optimization of parameters. The algorithms must have the capability to handle inputs from multiple domains that may contain various classes, and these classes may have a different kind of object in nature. The multiple objects in inputs can be handled by including localization while optimizing the parameters. In the open-world applications that are working in the real world, the input rate is a significant issue because of the unpredicted flow of input in terms of size and frequency. The open-space risk minimization is a crucial challenge for every algorithm to ac hive high accuracy while learning in the open-world. The system must include prior knowledge to adapt continuity in learning that can reduce learning efforts in the future.

Image processing is one of the binding domains of computer science, and there is plenty of work has been done in this field, although there is scope to extend the research in open-world machine learning . The world is towards automation in computer vision and image processing, such as driverless cars and humanless goods delivery systems introduced by many research organizations. The real-time activities in a dynamic environment can be handled if the system is interactive and functions end-to-end to recognize the multiple objects in open-space. The interactive models will help to scale real-time data handling capacity with multi-class objects, and they can be from different domains. Realistic results can be achieved if the system can deal with both empirical and open-space risks.  The use of past knowledge to recognize unseen objects in a dynamic environment will increase the accuracy of the system and provide more realistic results. Thus the knowledge base must be updated incrementally. The following challenges we observed in OWML for CV-IP tasks.

\begin{itemize}
    \item Open-space and empirical risk parameters are not optimized. Therefore, many models cannot adapt generalization or specialization.
    \item Most of the recommended methods have used limited training sampling; hence, the real-world impacts can not be determined accurately.
    \item Most of the recommended methods have not been employed with localization; hence, it is insufficient to address images with multiple objects.
    \item There is an absence of a mechanism for the minimization of open-space risk. The learning can be improved by employing a dictionary learning-based algorithm for open-set recognition.
    
\end{itemize}

\section{Open-world Machine Learning  in Natural Language Processing (NLP)} \label{Section 5}
Over the years, there has been enormous content generated on the web in the form of text. Social media is where billions of users create most of the text that can influence human beings and social sentiments in terms of thoughts, stories, expression, news, and daily life events. Social media is a crucial part of the current environment in terms of social and political perspectives. It can influence billions of people of the world positively or negatively by injecting synthetic views that can be already part of any plan. Therefore, analysis of social media content is vital to guide the world in a positive direction. Some work has been done on text data to analyze the text in different ways. open-world machine learning  can help us learn about the text in a dynamic environment. Text classifications and analysis of data is the utmost imperative entity for any organization. Standard text classification includes sentiment analysis, spam filtering, movie genre reviews, and document classification. The classification of tasks and work done towards these tasks are shown in Table \ref{taskwisenlp}. Further, These tasks are discussed in Section \ref{IL}, \ref{nlpdui}, \ref{KGandKB}, and \ref{nlpdnc}.

\begin{table}[htb]
\caption{Summarized Study of Task  Performed by Open-world Machine Learning in Neutral Language Processing}
\label{taskwisenlp}
\centering
\scriptsize
\begin{tabular}{c|c|c|c|c}
\hline
\multicolumn{1}{c|}{\textbf{Author(s)}}                             & \textbf{IL}      & \textbf{DUI}     & \textbf{KB\&KG}    & \textbf{DNC}     \\ \hline
G. Fei and B. Liu~\cite{fei2016breaking}               & -       & $\surd$       & -       & -       \\ \hline
L. Shu et al.~\cite{shu2017doc}                       & -       & $\surd$ & -       & -       \\ \hline
S. Prakhy et al.~\cite{prakhya2017open}               & -       & $\surd$ & -       & -       \\ \hline
X.  Guo et al.~\cite{guo2019multi}                  &  -       & $\surd$       & -       & -       \\ \hline
T. Doan and J. Kalita~\cite{doan2017overcoming}        & -       & $\surd$ & -       & -       \\ \hline
B. Shi and T. Weninge~\cite{shi2018open}               & -       & -       & $\surd$ & -       \\ \hline
S.  Mazumde et al.~\cite{mazumder2018towards}         & -       & -       & $\surd$ & -       \\ \hline
T.-E. Lin and H. Xu~\cite{lin2019deep}                 & -       & $\surd$ & -       & -       \\ \hline
H. Xu et al.~\cite{xu2019open}                        & -       & $\surd$ & -       & -       \\ \hline
N. Vedul et al.~\cite{vedula2019towards}              & -       & $\surd$ & -       & -       \\ \hline
T.-E. Lin and H. Xu~\cite{lin2019post}                 & -       & $\surd$ & -       & -       \\ \hline
G. Fei et al.~\cite{fei2016learning}                  & $\surd$ & -       & -       & -       \\ \hline
N. Vedula et al.~\cite{vedula2020automatic}           & -       & -       & -       & $\surd$ \\ \hline
Q. Wu et al.~\cite{wu2021towards}                     & -       & $\surd$ & -       & -       \\ \hline
Y. Wang et al.~\cite{wang2021caps}                    & -       & -       & $\surd$ & -       \\ \hline
\end{tabular}
\end{table}

\subsection{Incremental Learning (IL)} \label{IL}

Incremental learning is a Machine Learning (ML) method concerns expanding artificially intelligent systems that can continue to learn new tasks from novel input while retaining previously gained knowledge. Whenever a novel task(s) appears and changes, the training method occurs. The model keeps whatever has been learned according to the novel task(s) and old knowledge. The most notable distinction of incremental learning from conventional machine learning is that it does not lose previous knowledge. However, the training samples resemble it over time.

In~\cite{fei2016breaking}, authors proposed Center-based Similarity (CBS) method for open-world text recognition. It is a space learning method that can reduce open-space risk. The CBS is based on SVM. Center-based similarity space learning transforms each document space vector or feature vector, each feature in the center of the positive class document, and the feature vector of the document. At the same time, traditional classification directly uses training examples for trained binary text classifiers. CBS can learn multiple documents features vectors, separate for each document, and represents the center for multiple positive documents. Similarity value can be computed using multiple document similarity functions. The performance evaluations have been done on two publicly available datasets, 20-Newsgroup~\cite{lang20newsgroups,lang1995newsweeder}, and amazon customer reviews~\cite{mudambi2010research}. 

 In~\cite{fei2016learning}, authors extend their work and given a better system that can practice incremental learning in which the system can learn cumulatively. Whenever the system learned about new classes /unseen classes became more knowledgeable, just like humans do. They proposed a system with two specific abilities, continually detecting unknown classes and cumulatively adding the data of these new classes to the knowledge base without re-train the whole system. The proposed Center-based Similarity Space Learning SVM (CBS-SVM) was evaluated with two different datasets Amazon product reviews of 100 domains and 20-newsgroup~\cite{lang20newsgroups,lang1995newsweeder}. Classifying classes in the open-world uses the same unseen class rejection method based on threshold probabilities. The system used a similarity method to learn unseen/ new classes. It searched for sets of similar classes and learned to separate new classes. To learn a separate new class, it builds a binary classifier. After detecting or specifying a new class, updates the existing classifier to avoid confusion for the next unseen classes. The proposed method was evaluated by comparing the result with existing 1-vs-rest-SVM, 1-vs-set-linear, WSVM-linear, WSVM-RBF, PI-SVM-linear, PI-SVM-RBF, ExploratoryEM, CBS-SVM performed better than all existing methods with all different openness.

\subsection{Discovery of Unseen Instances (DUI)} \label{nlpdui}

open-world machine learning  has the significant importance of rejection of unseen classes; the accuracy of prediction of the known class must be justifiable. In~\cite{shu2017doc}, authors proposed Deep Open Classification (DOC) to identify new classes or tasks which may not belong to any training class. The ideal classifier should document both for which training class is available and the document for which training class is not available. This method is called open-world classification or open classification. 

Giving the training data set $D= \{(A_1,B_1), ( B_i, A_2), \ldots , ( A_n, B_n)\}$ Where $A_i$, is  $i^{th}$ document and $B_i$ = $\{l_1, l_2, \dots  l_m\}$ = $B$ is  $A_i$ class label. They build classifier $f(x)$ it can classify test instances A such that, A belongs to the training class $m$ as a seen class in B or discard it that means it is unseen class, and test instances do not belong to any of $m$ training class or any other seen class. DOC will build a multi-class classifier with the 1-vs-rest final layer of sigmoid in place of OpenMax~\cite{scheirer2011meta} to reduce open-space risk. DOC used the sigmoid function with Gaussian fitting to lighten the decision boundaries and reduces open-space risk. DOC used a Convolutional Neural Network (CNN) with a 1-vs-rest sigmoid layer and Gaussian fitting for classification. DOC Chose CNN because OpenMax uses CNN, and CNN performs well on the text. Doc has three layers for a different task. Layer 1: Embedded word (word vectors pre-trained from Google News that is Word2Vec)~\cite{mikolov2013efficient,mikolov2013distributed} in $x$ document into a dense vector. Layer 2: Perform convolution on layer 1 with the different filters with a variety of sizes. Layer3: A pooling layer selects a maximum value from the result of layer-2 and forms a K-dimension. To evaluate the performance experiment has been done on two datasets, 20 Newsgroups~\cite{lang20newsgroups} and 50-class reviews~\cite{chen2014mining}and compare results with CBS-SVM~\cite{fei2016breaking} and OpenMax. To extract the features, they convert the document into vectors using word2Vec~\cite{mikolov2013efficient,mikolov2013distributed} method. They used pre-trained from Google News vector~\cite{mikolov2013efficient} that consist of three million words and 300 dimensions word for word to vectors.

In~\cite{doan2017overcoming}, the authors proposed the Nearest Centroid Class (NCC) to detect unseen classes in open-world machine learning . It is an incremental learning method, which can take sets of closest neighbors of the centroid class. There are clusters for classes, and in a cluster, each class has minimum points. These are the membership points that are associated with clusters. Each class must have a minimum membership point to join the particular cluster. The class also represents the data point, and the center of the class is the data points.  New classes that have the nearest class center data point allow joining the cluster. To evaluate the performance of the algorithm experiment done on 20-newsgroups and amazon reviews datasets with different numbers of domains.  The prior algorithm performed better for some of the parameters for both datasets, but NCC's overall performance is significantly better.

 ChatBots can work in a dynamic open-world environment, but it is vital to recognize the user's intention.  Intent classification is a technique to distinguish the perseverance or intention by estimating the text language. It refers to an intent classification or intent identification. Nowadays, many institutions use text-based chat systems to solve their customers' queries without any human interactions. ChatBots must understand the unknown intentions of the user to work as a human being.
 
 In~\cite{prakhya2017open}, the authors proposed another CNN-based approach. It is based on feature extraction. To extract the features, they convert the document into a vector using word2Vec~\cite{mikolov2013efficient,mikolov2013distributed} method. To calculate the document vector, they used naïve methodology and estimated the cosine similarity among the mean of the document of the entire document vector. Deep learning models are used for open text classification with a modified Weibull layer as the final layer instead of the traditional SoftMax layer~\cite{goodfellow2016deep}. It is single-layer architecture, but the experiment has been done with different no of layers. To evaluate the performance of the proposed model, evaluation of proposed technique has been done on 20-Newsgroups~\cite{lang20newsgroups} and Amazon product reviews dataset and compared the results with existing methods. 
 
 In~\cite{lin2019deep}, the authors proposed two-stage methods for detection of unknown intent in the dialog system. To extract the feature of unknown intent, it uses Bidirectional Long Short-term Memory (BiLSTM) network using a margin loss. The LSTM network minimizes the variances of intra-class and maximizes the variances of inter-class intents. Glove word embedding used to create vectors and to distinguish the unknown intent local outlier factor LOF~\cite{breunig2000lof} has been used. The loss layer detects the known intents from deep discriminative features, while LOF detects unknown intents. The SNIPS~\cite{coucke2018snips} and ATIS~\cite{tur2010left} dataset has been used to evaluate the result of the proposed method. The performance of the method is compared with Maximum Softmax Probability (MSP)~\cite{hendrycks2016baseline}, DOC~\cite{shu2017doc}, DOC SoftMax, and LOF SoftMax.
 
 In~\cite{guo2019multi} authors proposed a Deep convolutional neural network (DCNN) which is cascade architecture that can continue to learn newer classes. The framework is an end-to-end  Open-world Recognition (OWR). To detect the instances from unknown classes, they proposed Multi-stage Deep Classifier Cascades (MDCC). It contains unique features for known classes and can distinguish the class as a known class at any stage of the process. Incremented leaf nodes can detect features of unknown classes and recognize newly added classes. It can learn new features of recently added classes without wounding existing features of known classes. The evolution of MDCC was done on the RF signal and Twitter dataset~\cite{guo2019multi,purohit2014emergency}. The experimental outcomes are compared with Local Novel Detector (LOD)~\cite{bodesheim2015local}, S-Forest~\cite{mu2017classification} and R-OpenMax~\cite{moore2015context}.

The e-commerce industry is growing and has become a significant part of the world economy. Product classification is one of the most important aspects of any e-commerce organization. The unpredicted or unknown search about the product is critical for these industries as different categories of products appear every day. The queries which are not predefined or known for the system can affect the reliability of the entire organization. In~\cite{xu2019open}, the authors proposed open-world learning (OWL) model Learning to Accept Classes (L2AC), which is based on meta-learning. L2AC maintains only dynamic known classes that allow novel classes to be added without retrained the model. In L2AC, each known class acts as a small set of the training example. The testing uses only Meta-classifier (using known and novel classes). The L2AC model has two primary mechanisms, ranker and meta-classifier. The ranker retrieves examples from known classes that are comparable or nearest to test examples. The meta-classifier is the core mechanism of L2AC, and it is a binary classifier that distinguishes the classes as known based on probability score or rejects otherwise. To evaluate the performance of L2AC outcomes compared with a different variant of DOC~\cite{shu2017doc} on the Amazon data set, the L2AC shows effectiveness for some parameters.

In~\cite{vedula2019towards}, the authors proposed a model  Towards Open Intent Discovery (TOP-ID) for open intent detection. It is a two-phase mechanism that predicts the intent for the statement and then tags the intent in the input statement. The model consists of a BiLSTM~\cite{schuster1997bidirectional} and Conditional Random Field (CRF) with the adversarial training method, and it increases robustness and performance through the domain. TOP-ID can detect a user's intent automatically in natural language. It does not need any prior knowledge for intent detection. The first part of TOP-ID detects existing open intent and then tags it into input words with action and objective. If there is no objective and action associated with detected intent, then it is tagged as none. To perform this task initially, convert the text into feature sequence by assembling character level representation, obtained by using a CNN with Glove word embedding~\cite{pennington2014glove}. To avoid combined word embedding effect on accuracy, TOP-ID used Highway Network~\cite{srivastava2015highway}. The second module of TOP-ID is the intent discovery framework. It takes adversarial inputs (close to the original) created by adding noise in data in the form of perturbations. The overall training has been done with both original and adversarial inputs. The attention mechanism is part of the intent discovery framework. There are multiple attention functions used that attend the information of the input sequences at different positions. Finally, the CRF predicts one of the three tags for the sequence of the words. To evaluate the TOP-ID, they create a dataset by collecting 75K questions with correct answers then annotating 25K quotations data  (three tags action, object, and none) with the amazon technique. The F1- Score of TOP-ID is significantly better than existing methods.

In~\cite{lin2019post}, the authors proposed a Softmax and Deep Novelty (SMDN)   detection model to detect unknown intents. The SMDN classifiers can be functional on any model without altering the architecture of the existing model. The model uses SoftMax that classifies by calculating the calibrated confidence score, and detects unknown intent by calculating decision boundary. The LOF~\cite{breunig2000lof} is used as an output layer to detect the unknown intent. To evaluate the performance outputs are compared with different variant of DOC~\cite{shu2017doc} on three SNIPS~\cite{coucke2018snips}, ATIS~\cite{tur2010left} and SwDA~\cite{jurafsky1997switchboard,shriberg1998can,stolcke2000dialogue} benchmark datasets.

In~\cite{wu2021towards}, the authors propose a Inductive Collaborative Filtering (IDCF) system, which provides inductive learning for user inputs while also ensuring sufficient expressiveness and adaptability. The IDCF uses two representation models to extract user-specific embeddings, that term meta latents. It factorizes a set of essential users' data matrices, followed by an attention technique that learns concealed graphs among essential users and queries users based on their past ranking habits. For query users, the inductive calculation of user-specific representations is enabled by the revealed associated graphs. IDCF standard version can decrease restoration loss to a similar level as vanilla matrix factorization technique under a slight circumstance. Empirically, IDCF offers actual close Root Mean Square Error (RMSE) to transductive Collaborative Filtering (CF) models. IDCF achieves improved results over the few-shot and unseen users compared to several inductive models on explicit feedback data Movielens-100K, Movielens-1M~\cite{harper2015movielens}, and Douban~\cite{yang2019douban}, and implicit feedback data Amazon-Beauty, and AmazonBooks~\cite{he2016ups,mcauley2015image}.

\subsection{Knowledge-graph \& Knowledge-base (KG \& KB)} \label{KGandKB}

The knowledge graph (KG) is one of the key methodologies for the online and offline world. The KG is helping in many important tasks such as Web search, entity linking, language processing, recommendation, and prediction. This method is also worked under the close-world assumptions as nodes are predefined. Very few research is available for open-world machine learning  through the graph completion method. The relation and triples~\cite{yue2020representation} are key components for knowledge graph completion methods. In~\cite{shi2018open} authors proposed a ConMask, which is a model for Knowledge Graph Completion (KGC) in the open-world. This model learns embedding of any entity by its name, description given in text fields and identifies unknown classes of entities to the knowledge graph. ConMask used relation depending content masking to extract relevant chunks and reduce the noisy text description. After extracting relevant chunks, train the model with fully connected CNN to concur chunks with entities in a knowledge graph. Knowledge graphs can be representing as $(H_d,R_e,T_a)$ where $H_d$ is head, $R_e$ is relation between Head $H_d$ and some tail entity $T_a$. The generalize incomplete graph $G = (e,\  r,\  t)$, where $e$ = entity set , $r$= relation set and $t$= tuple set. This graph can be complete by finding missing tuples $t'$. $t'= \{ \langle H_d, R_e, T_a \rangle | H_d \in e, R_e \in r, T_a \in e, \langle H_d, R_e, T_a \rangle \in t \}$ in the incomplete KG. The model consists of three modules, relationship content masking, target fusion, and target entity resolution. The first module indicates the words relevant to the task. The second module extracts target entities embedding. The last module picked the target entities based on the similarity score between target entities. The last module of the model furthermore extracts entity embedding and textual features.

As the world is going towards automation, ChatBots systems are becoming popular for customer common query solutions.  Every system that takes input as a text to elucidate or respond to a User’s inquiry needs to understand the query. Existing systems are working with a limited environment, which means they can answer the only query for data available in the Knowledge Base (KBs). Such systems have limitations; they cannot work in a dynamic environment because they cannot learn new knowledge. Although KBs have extensive data from appropriate sources still much information has missing from them. Many techniques have been proposed till now to complete this missing information. These methods are termed KB completion, but all the methods worked under closed-world assumptions. KBs has a limitation, and it cannot work in an open-world environment.

To address this kind of issue, in~\cite{mazumder2018towards}, the authors proposed Open-world Knowledge Base Completion (OKBC) and Lifelong Interactive Learning and Inference (LILI) technique for ChatBots to acquire knowledge in the dialogue process. This knowledge learning engine allows ChatBots to gain knowledge throughout the conversation and make it further interactive.  It is based on the theory of continual learning, where ChatBots become more knowledgeable with time as they learn continually after every conversation. Lifelong interactive learning and inference analyze the query and add it to KB if it does not exist. LILI, formulate an inference strategy, learn interaction behaviors, leverage the acquired knowledge, and continuously repeated this to learn new knowledge. The evolution of LILI has been done on two benchmark datasets Freebase FB15k~\cite{bordes2013translating} and WordNet~\cite{miller1995wordnet} and compare results for known, unknown, and overall classes. The result of LILI is effective for both predictive eminence and strategy formulation capability.

In~\cite{wang2021caps}, The authors suggest a capsule network-based approach Caps‑ OWKG that leverages context to describe relationships and objects in the open-world knowledge graph. The proposed Caps‑OWKG consists of triplets that are the basic unit of the system. In addition, the capsule network conducts extraction of features, judgment on triplets, text synthesis, and fusion analysis. When computing triplets, the Caps‑OWKG technique has the benefit of providing a stronger connection between items and relationships. These interpretations are also refined; thus, the Caps‑OWKG model may be considered a dynamic embedding exploration that accurately represents the triplet. The existing known techniques such as ConMask~\cite{shi2018open}, TransE-OWE~\cite{shah2019open}, and DKRL~\cite{xie2016representation} are used to compare the performance of  Caps-OWKG on the two benchmark datasets FB15k-237~\cite{toutanova2015observed} and DBPedia50k~\cite{shah2019open}, achieving better outcomes than existing techniques.

\subsection{Detection of Novel Classes (DNC)} \label{nlpdnc}

The existing research finds only new intent in the available domain, but the novel domain is introduced incrementally as data increases. The novel domain must be found to make the system fully automated and reduce the limitations of the system. In~\cite{vedula2020automatic} authors proposed Automatic Discovery of Novel Intents and domains (ADVIN) to discover novel domains and intents of text from unlabeled data. ADVIN works in three stages: discovering the novel domains and intent from extensive unlabeled data, knowledge transfer, and linking related intents to corresponding novel domains. To identify the instances of novel intents ADVIN, used BERT~\cite{devlin2018bert} based multi-class classifiers. The DOC~\cite{shu2017doc} is used for distinguishing unseen intents. In the second stage that discovers the categories of newly discovered intents, it uses a hierarchical clustering method to transfer knowledge. Finally, by linking novel intents into novel domains, ADVIN used clusters of seen classes as ideal clusters and knowledge transfer modules to represent clusters. To evaluate the performance of ADVIN results are compared with DOC, IntentCapsNet~\cite{yue2020representation}, LOF-LMCL~\cite{doan2017overcoming} and different combinations of ADVIN and DOC on four benchmark datasets, SNIPS~\cite{coucke2018snips}, ATIS~\cite{tur2010left}, Facebooks’ Task-oriented Semantic Parsing (FTOP)~\cite{gupta2018semantic} and dataset from a commercial voice assistant, Internal NLU. The overall performance of ADVIN is significantly better.

Table \ref{summmaryNLP} shows a summarized illustration of literature on open-world machine learning in Natural language processing.  It shows the used or recommended methodology, datasets employed for evaluation, and proposed results by the authors.

\begin{scriptsize}
\begin{longtable}{p{1.5cm}|p{2.0cm} |p{2.8cm}|p{6.0cm}}
\caption{Summary of the Proposed Approaches for Natural Language Processing in open-world machine learning }
\label{summmaryNLP}

\\ \hline
\textbf{Author} & \textbf{Proposed or Used Methodology} & \textbf{Datastes}  & \textbf{Reported Results} \\ \hline
G. Fei and B. Liu~\cite{fei2016breaking}        & Center-based Similarity support vector machine (CBS-SV)                           & 20-newsgroup and Amazon reviews           & Accuracy 45 to 87.3\% for 25\% to 100\% openness (Maximum with Amazon reviews 10 Domains)                 \\ \hline
G. Fei et al.~\cite{fei2016learning}           & Cumulative Learning (usine CBS-SVM)                          & Amazon product reviews and 20-newsgroup         & Micro F1-score 66.2 to 83.5\% for openness of 33\% to 100\% (Maximum with 20-newsgroup )                 \\ \hline
L. Shu et al.~\cite{shu2017doc}                & Deep Open Classification (DOC)                          & 20 Newsgroups and Amazon reviews (50-class reviews)          & Micro F1-score 82.3 to 92.6\% for 25\% to 100\% openness (Maximum with 20 Newsgroups)                  \\ \hline
S. Prakhy et al.~\cite{prakhya2017open}        & Convolutional Neural Networks (CNNs)                          & 20-newsgroup and Amazon reviews           & F1-score 79.7 to 82.1\%  for 25\% to 100\% openness (Maximum with Amazon reviews 10 Domains)                \\ \hline
T. Doan and J. Kalita~\cite{doan2017overcoming} & Nearest Centroid Class (NCC)                           & 20 newsgroups and Amazon reviews     & Accuracy 20 to 82\% for 0 to 50 domains (with Amazon revies)                  \\ \hline
X.  Guo et al.~\cite{guo2019multi}                       & Multi-stage Deep Classifier Cascades (MDCC)                       & RF signal Datasets,Twitter dataset,                          & EN-Accuracy 60.45\% (Maximum with RF Signal) F1-score 75\% (Maximum with RF Signal)                            \\ \hline
B. Shi and T. Weninge~\cite{shi2018open}        & Content Masking (ConMask)                           & DBPedia50k and DBPedia500k          & Mean Rank 90 and Mean Reciprocal Rank 35.0 (for head) 
Mean Rank 16 and Mean Reciprocal Rank 61.0 (for trail) both are maximum with DBPedia50k \\ \hline
S.  Mazumde et al.~\cite{mazumder2018towards}  & lifelong inter-
active learning and inference (LILI)                          & Freebase (FB15k1) and WordNet         & Avg. F1-score 63.43\% (Maximum with FB15k1) and Avg. MCC 39.39\% (Maximum with WordNet)                \\ \hline
T.-E. Lin and H. Xu~\cite{lin2019deep}          & Bidirectional
long short-term memory (BiLSTM)                           & SNIPS and ATIS         & F1-score 78.8 to 79.2\% for 25\& to 75\% openness (Maximum with SNIPS)                  \\ \hline
H. Xu et al.~\cite{xu2019open}                 & Learning to
Accept Classes (L2AC)                          & Amazon Datasets         & Micro F1-score 84.68 to 93.19 \% for 25\% to 75\% openness                 \\ \hline
N. Vedul et al.~\cite{vedula2019towards}       & Towards Open Intent Discovery (TOP-ID)                          & 25k real-life utterances (Created dataset)          & F1-score 91\% (Maximum among all the versions of TOP-ID)                  \\ \hline
T.-E. Lin and H. Xu~\cite{lin2019post}          & SofterMax and deep
novelty detection (SMDN)                           & SNIPS, ATIS, and SwDA           & Macro F1-score 71.1 to 79.8\% for 25\% to 75\% openness (Maximum with SNIPS)                 \\ \hline
N. Vedula et al.~\cite{vedula2020automatic}    & Automatic Discovery of Novel Intents (ADVIN)                         & SNIPS, ATIS, FTOP,  and Internal NLU Dataset        & F1-score  92\% (for discovery of unseen instances) NMI 83\% Purity  92\% F1-score  78.0\% (for discovery of unseen classes)                \\ \hline
Q. Wu et al.~\cite{wu2021towards}              & Inductive Collaborative Filtering (IDCF)
model                          & Douban, Movielens-100K, Movielens-1M, Amazon-Books, and Amazon-Beauty        & AUC 94.4\% (Maximum with Amazon-Books) and  Normalized discounted cumulative gain (NDGC) 95.5\% (Maximum with Douban)           \\ \hline
Y. Wang et al.~\cite{wang2021caps}             & capsule-network for open‑world knowledge
graph (Caps-OWKG)                           & DBPedia50k and FB15k-237-
OWE         & Tail prediction 64.8\% (Maximum with DBPedia50k)                 \\ \hline
\end{longtable}
\end{scriptsize}

\subsection{Available Software Packages and Implementations}

In this section, we provided a link for some software packages which contains the various implementation of various model of open-world machine learning  in natural language processing (Table \ref{softwareNLP}). These are the models commonly used in various frameworks of open-world machine learning .

The available software packages can be used to further improve learning in the open-world for natural language processing. The Open-set Deep Networks (OSDN) can be used for open-set reorganization, Deep Open Classification (DOC) for unseen class identification, and  Content Masking (ConMask) for identification of unseen entities in the knowledge graph. There are two packages word to vector (Word2Vec) and  Global Vectors (Glove), for input word embedding.

\begin{table}[htb]
\caption{Available Software Packages and Implementation}
\label{softwareNLP} 
\centering
\scriptsize
\begin{tabular}{p{2.9cm}|p{2.2cm}|p{7.5cm}}
\hline 
\textbf{Author} & \textbf{Model} & \textbf{Link}                                                                         \\ \hline 
{P. Moore and H. Van Pham~\cite{moore2015context}}   & OSDN                        & https://github.com/abhijitbendale/OSDN                                        \\ \hline
{L. Shu et al.~\cite{shu2017doc}}  & DOC                         & https://github.com/leishu02/EMNLP2017\_DOC                                    \\ \hline
{B. Shi and T. Weninger~\cite{shi2018open}}  & ConMask                     & https://github.com/bxshi/ConMask.                                             \\ \hline
T. Mikolov et al.{~\cite{mikolov2013distributed}} & Word2Vec                    & https://code.google.com/archive/p/word2vec                                    \\ \hline
{J. Pennington et al.~\cite{pennington2014glove}} & Glove                       & https://nlp.stanford.edu/projects/glove/                                      \\ \hline
{X. Guo et al.~\cite{guo2019multi}}        & MDCC           & https://github.com/xguo7/MDCC-for-open-world-recognition      \\ \hline
{Y. Kim et al.~\cite{kim2014convolutional}} & CNN Text   Classification   & https://github.com/dennybritz/cnn-text-classification-tf                      \\ \hline
{Y. Kim et al.~\cite{kim2014convolutional}} & CNN Sentence Classification & https://github.com/alexander-rakhlin/CNN-for-Sentence-Classification-in-Keras \\ \hline
\end{tabular}
\end{table}

\subsection{Discussion}
There are very few works that have been done in open-world machine learning  for natural language processing. The semantic similarity in the text is hard to address while learning new knowledge about the text at run time, especially when there is no training set available for such data. To achieve valuable outcomes from any framework or an algorithm, it must distinguish the semantic similarities in text. Therefore there is a need for large scale knowledge-base to learn the hierarchical structure of text words with meanings. Openness is a significant issue as we have analyzed that the accuracy has been reduced whenever the openness increased. There is a need for frameworks that can deal with dynamic values of openness and provide high accuracy with maximum openness. The automated dialog-based system that is quite popular nowadays needs a mechanism to process informal conversation in real-time.

The automated ChatBot systems and text and voice-based assistance devices are increasing rapidly in this decade, and it further improves the world of automation. To increase the accuracy of such a system, the open-space risk and distinguish the semantic similarities is one of the significant aspects of neural language processing in open-world settings. The cumulative and incremental model can help to address such issues. The system will produce more realistic outputs when it adapts scalability in input with maximum openness as the real-world inputs are unstructured. In this section, we have discussed various problems associated with NLP in open-worlds learning and proposed solutions by various authors that can help improve a text-based application working in a real-world domain and dynamic environment.  The following challenges we observed in OWML for NLP tasks.

\begin{itemize}
    \item Stable performance can be achieved to identify unseen instances only if the threshold value is within a reasonable range.
    \item The recommended methods show superior outcomes for sample instances. The accuracy of many of the proposed systems decreases if the number of seen classes is low. 
    \item There is a need for improvements to use these prototypes for real-time systems, as extensive experiments with large-scale datasets are missing.
    \item Only a few methods are employed with cumulative learning.
    \item In NLP, many of the recommended models suffer when distinguishing unseen intent from seen intents where semantic meanings are similar.
\end{itemize}

\section{Datasets Used in Open-world Machine Learning}
Most of the researchers employed benchmark datasets to evaluate the performance of their proposed algorithms.  Some of the researchers built their datasets or altered the existing dataset and evaluated the methods. In this section, we discuss some of the datasets primarily used in both the domain of open-world machine learning. 

\begin{figure}[htb]
    \centering
    \scriptsize
   \includegraphics[height = 2.8 in, width = 4.0 in]{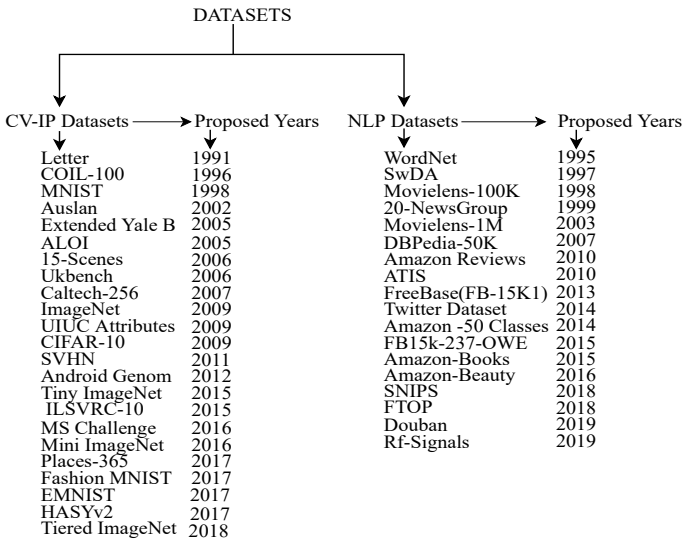}
    \caption{Classification of Datasets Used in CV-IP and NLP with Proposed Years}
    \label{datasetyears}
\end{figure}

Figure \ref{datasetyears} shows the classification of datasets for CV-IP and NLP with their proposed years. Next we discussed publicly available datasets that are used in OWML.

\textbf{Caltech-256~\cite{griffin2007caltech}:} $caltech-256$ has set of $256$ categories of object and the total $30607$ images in this dataset. Each category contains minimum $80$ and maximum $827$ images, these categories are further labeled with three tags on the basis of image quality. the labels are \textit{good},\textit{ bad} and \textit{none} (out of the category). the \textit{good} indicates clear vision and \textit{bad} indicates clutters or artistic example where \textit{none} indicates the image does not belong to the particular category.

\textbf{MNIST~\cite{lecun1998mnist,lecun1998gradient}:} Modified National Institute of Standard and Technology, wildly known as $MNIST$ is a handwriting dataset. It is a modified version of NIST. MNIST is used in optical character reorganization and is also used as a test case in pattern recognition and machine learning. We have analyzed, MNIST has become a standard for testing machine learning algorithms. 
There are $60000$ training images; some may use for validation and $10000$ images for testing purposes. All the digits are black and white and normalized in seize, the center intensity with $28*28$ pixels; thus, the dimension of the image is $28*28=784$, and each element is a binary. The MNIST has tested for almost all the benchmark baseline algorithms and well-known Fields of classification such as linear classification, convolution neural networks, simple neural networks, K-Nearest neighbors, support vector machines (SVMs), boosted stamps, and nonlinear classification.

\textbf{Fashion-MNIST~\cite{xiao2017Fashion}}: The Fashion-MNIST is a dataset of Zalando's article images which containing a training set of 60,000 images, and a test set of 10,000 images. Each image is a 28x28 pixel and grayscale image related to a label from 10  different classes. Zalando aims for Fashion-MNIST to serve as a substitution for the original MNIST dataset, which comprises many handwritten digits, for benchmarking artificial intelligence and machine learning algorithms. 

\textbf{ImageNet~\cite{deng2009imagenet}:} $ImageNet$ is a large scale ontological dataset of visual objects. The structure of $ImageNet$ inspired from $WordNet$ dataset thus it constructs on backbone of $WordNet$. $ImageNet$ has $80000$ synsets of $WordNet$ with around $1000$ full resolution cleaned images and its updating continuously. The basic $ImageNet$ contains $ 3.2$ million images with $12$ sub-trees and $5247 synsets$. $ImageNet$ is hierarchical dataset like $WordNet$ which contains the synonym's of world in tree structure.
The $12$ sub-trees consist of the following categories: bird, reptile, vehicles, musical instruments, tools, fruits, mammals, fish, amphibians, geological formulations, furniture, and flower with $5247$ synsets. The continual updating of this data aimed at 50 million images in a hierarchical structure. The evolution with various baseline methods showed that the $ImageNet$ has a $99.7$ percent average precision rate.

\textbf{Tiny-ImageNet~\cite{le2015tiny}:} $Tiny-ImageNet$ is a collection of $100000$  images that are retrieved from internet. The resolution of all these images is $32*32$ pixels and $64*64$. Tiny-ImageNet has 200 categories of images, of which $100,000$ images for training, $10000$ for validation, and $10000$ images are reserved for testing.  Images are collected by sending all search words in $WordNet$ to the image in the search engine. it is a successful dataset tested on application-specific algorithms because of a high level of noise and low resolution. $Tiny-ImageNet$ is suitable for general-purpose algorithms. It also contains synsets of high quality with an average resolution of $400*350$.   

\textbf{CIFAR-10~\cite{krizhevsky2009learning}:} The CIFAR-10 data set was developed by the Canadian Institute for Advanced Research. It contains $10$ categories (dog, frog, automobile, bird, horse, ship, truck, airplane, cat, and deer) of images, as it is a subset of CIFAR-100, which consists of $100$  categories of images. Total $60000$ color images are in CIFAR-10 with the resolutions of $32*32$ pixels, and every class has 600 images. The dataset is divided into training and test sets, which consist of $50000$ and $10000$ images, respectively. The entire dataset is divided into batches, $5$ batches for training and $1$ for testing. The testing batch has $1000$ random images, and the rest of the images randomly contain by training batches.

\textbf{SVHN~\cite{netzer2011reading}:} The Street View House Number (SVHN) contains $600000$ labeled digits that are cropped from actual street view images. The initial goal of this dataset is to identify house numbers from original street view images. There are two types of images one is whole numbers, and another is cropped digits. The whole numbers contain high-resolution full-size original images with character-level bounding boxes for the house number. The cropped digits are character-level ground truth, and all these digits are resized with the resolution of $32*32$ pixels. The SVHN is further divided for training and testing, and there are $73257$ digits images for training and $26032$ digits images for testing. The rest of the images also reserve as extra for training. 

\textbf{20-NewsGroup~\cite{lang20newsgroups,lang1995newsweeder}:} It is a collection of near about  $20000$ new documents which are collected from different newsgroups. It is one of the most popular datasets for the application that is based on text classification in machine learning. All the newsgroups are different, but some of the new groups are related to each other. Generally, $90$ percent of documents are used for training and $10$ percent for testing. 20-NewsGroup is publicly available in different forms, and the original dataset is not sorted but later on is sorted by date. The headers and duplicate data are also removed in this version. The latest version of this data is available with $18828$ documents with only "From" and "Subject" headers.

\textbf{Amazon Product Reviews~\cite{mudambi2010research}:} Amazon.com is one of the most successful e-commerce web site across the globe since it emerges. The amazon product review dataset contains $5.8$ million reviews, written by $2.14$ million for $6.7$ million products from $9600$ different categories when extracted from these reviews. The dataset has $8$ different headers such as Product ID, Reviewer ID, Rating, Date, Review Title, Review Body, Number of Helpful Feedbacks, and  Number of Feedbacks. It can be used for feature identification and construction of both reviewer and reviews, and the features can be Review Centric or  Product-Centric. The amazon product reviews.

\textbf{50-Class Reviews~\cite{chen2014mining}:} The 50-Class review dataset is a collection of reviews, and there are $50$ different categories of products. The data set has two versions, one has reviews of  $50$ different electronic items, and the other has $50$ different non-electronic items. There are $1000$ reviews for each product or domain.

\textbf{WordNet~\cite{miller1995wordnet}:} WordNet is a multi-language (Approx 200 languages) lexical dataset of semantic relationships among words, including meronyms, synonyms, and hyponyms. Some synsets contain synonyms in a group with short definitions and examples. The WordNet is a popular dataset for text analysis applications in artificial intelligence and machine learning. Initially, it has created for the English language only; later on, it extended for other languages, and updating is continuous to add a new language in WordNet. The WordNet contains approx $175979$ words which are organized in $175979$ synsets, and there is a $207016$ pairs which are word-sense pair. All synsets are connected with semantic relations.

\textbf{SwDa~\cite{jurafsky1997switchboard,shriberg1998can,stolcke2000dialogue}:} The Switchboard Dialog Act Corpus (SwDA) covers the SwDA-1 Telephone Speech Corpus, and some tags recapitulate semantic, syntactic, and pragmatic information about the related turn.

Some publicly available datasets with their repository link are shown in Table \ref{datasets}.

\begin{table}[htb]
\caption{Publicly Available Benchmark Datasets Repositories}
\centering
\scriptsize
\label{datasets}
\begin{tabular}{p{3.0cm} |p{8.8cm}}
\hline 
\textbf{Dataset}                & \textbf{Link}                                                                         \\ \hline
Caltech-256{~\cite{griffin2007caltech}}            & http://www.vision.caltech.edu/Image\_Datasets/Caltech256                             \\ \hline
MNIST{~\cite{lecun1998mnist,lecun1998gradient}}                   & http://yann.lecun.com/exdb/mnist                                                     \\ \hline
Extended Yale B{~\cite{lee2005acquiring}}                   & http://vision.ucsd.edu/~leekc/ExtYaleDatabase/ExtYaleB.html                                                     \\ \hline
ALOI{~\cite{geusebroek2005amsterdam}}                   & https://aloi.science.uva.nl                                                     \\ \hline
UIUC Attributes {~\cite{farhadi2009describing}}                   & https://vision.cs.uiuc.edu/attributes                                                     \\ \hline
Mini ImageNet {~\cite{vinyals2016matching}}                   & https://cseweb.ucsd.edu/~weijian/static/datasets/mini-ImageNet                                                    \\ \hline

Fashion-MNIST{~\cite{xiao2017Fashion}}           & https://github.com/zalandoresearch/fashion-mnist/tree/master/data                     \\ \hline
HASYv2{~\cite{thoma2017hasyv2}}                   & https://https://zenodo.org/record/259444/files/HASYv2.tar.bz2?download=1                                                  \\ \hline

ImageNet{~\cite{deng2009imagenet}}                & http://image-net.org/download                                                         \\ \hline
Tiny-ImageNet{~\cite{le2015tiny}}           & http://cs231n.stanford.edu/tiny-imagenet-200.zip                                      \\ \hline
CIFAR-10{~\cite{krizhevsky2009learning}}                & https://www.cs.toronto.edu/$\sim$kriz/cifar.html                                      \\ \hline
RF Signal Dataset~\cite{guo2019multi}              & https://github.com/xguo7/MDCC-for-open-world-recognition
                                \\ \hline
Twitter Dataset~\cite{purohit2014emergency}              & https://github.com/xguo7/MDCC-for-open-world-recognition
                                \\ \hline
SVHN{~\cite{netzer2011reading}}                    & http://ufldl.stanford.edu/housenumbers                                               \\ \hline
20-NewsGroup{~\cite{lang20newsgroups,lang1995newsweeder}}   & http://qwone.com/$\sim$jason/20Newsgroups                                            \\ \hline
Amazon Product Reviews{~\cite{mudambi2010research}} & https://jmcauley.ucsd.edu/data/amazon                                                \\ \hline
WordNet{~\cite{miller1995wordnet}}                & https://wordnet.princeton.edu/download                                                \\ \hline
SwDa{~\cite{jurafsky1997switchboard,shriberg1998can,stolcke2000dialogue}}           & https://web.stanford.edu/$\sim$jurafsky/ws97/                                         \\ \hline
ATIS{~\cite{tur2010left}}                    & https://rasa.com/docs/rasa/nlu-training-data/\#json-format                            \\ \hline
FB15k{~\cite{bordes2013translating}}                  & https://www.microsoft.com/en-us/download/ confirmation.aspx?id=52312                   \\ \hline
DBpedia{~\cite{auer2007dbpedia}}                 & https://wiki.dbpedia.org/datasets                                                     \\ \hline
EMNIST{~\cite{cohen2017emnist}}                  & https://www.nist.gov/itl/products-and-services/emnist-dataset                         \\ \hline
Auslan{~\cite{kadous2002temporal}}                 & https://archive.ics.uci.edu/ml/datasets/Australian +Sign+Language+signs+(High+Quality) \\ \hline
Ukbench{~\cite{nister2006scalable}}                & https://archive.org/download/ukbench                                                  \\ \hline
Places-2 {~\cite{zhou2017places}}              & http://places2.csail.mit.edu/download.html                                            \\ \hline
\end{tabular}
\end{table}

\section{Baseline Algorithms Used in Open-world Machine Learning }
Some methods and algorithms are used in open-world machine learning  that are standard or base concepts to practice open-world learning. 

\subsection{Center-Based Similarity (CBS)~\cite{fei2016breaking}}
CBS is a classification method that classifies the data points into seen and unseen classes. It works on center-based similarity space learning technique. The CBS learns news classes incrementally and use 1-vs-rest layer to classify unseen classes~\cite{fei2016breaking}. The 1-vs-rest is one of key concept in open-world machine learning  to discover unseen classes. \newline Let us assume there is new class $l_{X+1}$, for learning it need a model $M_X$. Model $M_X$ consist set of $X$ 1-vs-rest binary classifiers $M_X = (m_1, m_2,... m_X)$, for the previous $X$ classes there is training dataset $D^{p_r} = (D_1, D_2,...D_X)$ ($p_r$  = previous) and corresponding labels are $S^X = (l_1, l_2,...l_X)$. Here each $M_i$ builds a binary classifier to identify $l_i$, when new dataset $D_{X+1}$ arrives for class $l_{X+1}$. the entire system functions for two task, to update $M_X$ and build new $M_{X+1}$ model to classifies all available instances in existing class $S^{X+1} = (l_1, l_2, ... l_X, l_{X+1})$  and recognize the $U_s$ unseen classes. 

\noindent \newline \textbf{Step 1:}search a \textit{set of classes} $S_c$ that are comparable to new class $l_{X+1}$,
\newline \textbf{Step 2:} learning for Isolate the new class $l_{X+1}$ and the previous classes in $S_c$.

\noindent \newline \textbf{Step 3:} $M_X = (m_1, m_2,... m_X)$ to classify instances in $D_{X+1}$ , the similarity between old classes $(l_1, l_2, ...l_X)$ and new class $l_{X+1}$ can be computed by using each of 1-vs-rest binary classifier $m_i$.

In  next step, new class $l_{X+1}$ separated and now for $S_c$ there is two task,
\newline \textbf{Step 4:} Build $M_{X+1}$ new classifier for $l_{X+1}$.
\newline \textbf{Step 5:} Update existing classifier as the classes  which are in  $S_c$.

\subsection{Incremental Class Learning~\cite{fei2016breaking}}

Incremental learning is encouraged by the thought of the human learning process. It learns most of the knowledge by an experience like humans do. It learns new knowledge by the time instead of finding existing knowledge.

Let us assume we have  Classification model $M_X = (m_1, m_2,... m_X)$ as input  and  $D^{p_r} = (D_1, D_2,...D_X)$ is previous dataset. The new dataset is $D_{X+1}$ and $\lambda_s$ is Similarity Threshold. we need classification model  $M_{X+1} = (m_1, m_2, ....m_X, m_{X+1})$ to learn incrementally using previous data. To obtain this the following steps 
need to be executed.

\noindent\textbf{Step 1:} Initialize $S_c$ to empty set.\newline
\textbf{Step 2:} Initialize the count and record total instances in $D_{X+1}$ (positive classified by $m_i$).\newline 
\textbf{Sept 3 :} Use $m_i$ and classify each instances in $D_{X+1}$ and record total positive instances classified by $m_i$. \newline
\textbf{Step 4:} check whether there are disproportionate instances in  $D_{X+1}$ as positive by $m_i$ to reduce class $l_i$. as resemblance  to class $l_{X+1}$. The ${{\lambda }_s}$ is threshold which regulate how many instances in $D_{X+1}$  should be classified $l_i$  before considering as analogous to $l_{X+1}$.\newline 
\textbf{Step 5:} build novel classifier $M_{X+1}$.

\subsection{Nearest Class Mean (NCM)~\cite{ristin2014incremental,mensink2013distance}}
The nearest class mean (NCM) is generally used for large-scale image classification. Two methods were used in most Research for large-scale image classification, k-nearest neighbor (K-NN) and nearest class mean (NCM), the nearest class mean (NCM) is more flexible than K-NN. The NCM characterizes classes by their mean feature vectors of its components.

Let us assume we have image $P$, which is represented in $D$-dimension with the feature vector.

$\vec{f}\in \mathbb{R}^{D}$.

\textbf{Step 1:} compute the class centroid  $c_a$ for each class $a\in A$.
\begin{equation}
    c_a = \frac{1}{P_a}\sum_{i\in P_k}\vec{f_i}
\end{equation}

where $P_a$ is the set of images label with class $a$ and the set of centroid (for each class) is $C= \{c_a\}$ and it has  cardinality $\left | C \right | = \left | A \right |$

\textbf{Step 2:} The classifier of  nearest class mean to classify an image $P$ will search closest centroid in feature space.

\begin{equation}
    a^{*}(P) = \overset{argmni}{a\in  A}\left \| \vec{f}-c_a \right \|^{2}
\end{equation}

where $\vec{f}$ is the feature vector of $P$

\subsection{1-vs-rest~\cite{rifkin2004defense}}
 Classical machine learning uses different functions as output for multi-class classification. However, These functions can not reject unknown classes. There is a need to normalized these functions for each class across the training classes to achieve rejection capability in the output mechanism.

The\textit{ 1-vs-rest} is one the method which provide rejection capability. Les us assume there is 1-vs-rest method with with $s$ sigmoid functions, where  $s$ is known objects. We have $i^{th}$ sigmoid function for $p_i$ class. The 1-vs-rest distinguish the classes as positive and negative for $p_i$ class such that $q=p_i$  is positive class and rest of all $q \neq p_i$  classes are negative. The loss can be calculated as log loss for all $s$ sigmoid functions for the training data.

\begin{equation}
  loss= \sum_{i=1}^{s}\sum_{j=1}^{n} - \mathbb{I}(q_j = p_i)log p_b(q_j = p_i)- \mathbb{I}(q_j \neq p_i)(1-log p_b(q_j = p_i))
\end{equation}

Where $\mathbb{I}$ is the Indicator function, j= 1 to $n$ ($n$ = Number of instances) and probability output of $s$ sigmoids for $j^{th}$ input of $i^{th}$ dimension of  $r$ is  $p(y_j = x_i) = sigmoid(r_{j,i})$ reject unseen classes such that

\begin{equation}
    \hat{q} =\begin{Bmatrix}
reject, \  if sigmoid (t_i )< k_i ,\   \forall \ p_i\ \in \ q_i\\ 
argmax_{p_i \in q} \  sigmoid(t_i), \  Otherwise 
\end{Bmatrix}
\end{equation}

Where $k_i$  is threshold which belongs to  $p_i$ , is probability $p_b$ is less than the threshold the input will be reject. 1-vs-Rest predicts class which has highest probability.

These are the benchmark algorithms that have been used in both computer vision and natural language processing to elucidate problems in open-world settings. There are several frameworks and models suggested which are centered on these algorithms or used these algorithms.    
The Center-Based Similarity (CBS) does not support the artificial neural network; thus the some of the authors used its extension or modified versions~\cite{alpaydin2020introduction,fei2016breaking}. The Nearest Class Mean (NCM) has been used as the baseline in ~\cite{bendale2016towards,de2016online}. The classical model of NCM  considers examples of training a novel 1-vs-rest classifier for individual supplementary classes. Hence in the case of large-scale datasets and multiple classes, it becomes a burden for classifiers. Earlier trained classifiers will also need to be restructured to increase their performance; thus, the extension of NCM such as Nearest Non-Outlier (NNO)~\cite{bendale2015towards}  implemented, which gives better accuracy. The Nearest Centroid Class (NCC)~\cite{doan2017overcoming} also inspired by NCM. In ~\cite{de2016online,shu2018unseen,doan2017overcoming,mazumder2018towards} the concept of incremental learning has been used to recognise objects in open-world.

 Identification of unknown data has a significant impact on open-world machine learning . To adapt the capacity of rejection of unknown data, many authors use the 1-vs-rest~\cite{rifkin2004defense} method. Some authors have used the 1-vs-rest method to identify unknown data~\cite{scheirer2012toward,shu2017doc,vedula2020automatic} and some researchers used this method as part of their framework to distinguish the known and unknown data~\cite{shu2018unseen,xu2019open}.
 As we have seen in literature, the baseline approaches are still generating auspicious outputs with current expertise such as Convolutional Neural Network (CNN) and Deep Neural Network (DNN) frameworks. We also perceive that the few modified versions of these algorithms reinforce the model in open-world machine learning .

\section{Related Areas}
Some related areas that are closely associated with open-world machine learning  are mentioned and discussed briefly in this section. 

\subsection{Transfer Learning}

Conventional Machine Learning (ML) techniques perform predictions on the expected data by applying analytical principles trained on previously accumulated unlabeled or labeled training examples~\cite{yin2006efficient, kuncheva2007classifier, baralis2007lazy}. 
Analysis on transfer learning has brought attention since 1995 in several titles:  knowledge transfer, learning to learn, multitask learning, knowledge consolidation, inductive transfer, knowledge-based inductive bias, context-sensitive learning, cumulative learning~\cite{thrun1998learning,thorisson2019cumulative}, and multitask learning framework ~\cite{liu2008semisupervised}.
Transfer learning involves interpreting data for a reference task to provide a productive basis for a new task. Transfer learning is often applied to specific data sets,  which have some labeled value. For example, an actual demonstrative prototype of one virus would have a significant advantage to developing a distinguishing prototype for another virus, for which fewer training samples are available. While all learning involves generalization across all queries, transfer learning illustrates the transfer of information across comparable but non-consistent fields, tasks, and distributions. In distinction, the unlabeled data does not require to be obtained from a similar task in the transfer learning framework. In the prior decade, there has been substantial development in improving cross-task transfer utilizing both discriminative and generative strategies in a broad category of frames. 

\subsection{Active Learning}
\begin{figure}[htb]
    \centering
   \includegraphics[height = 1.5 in, width = 3.5 in]{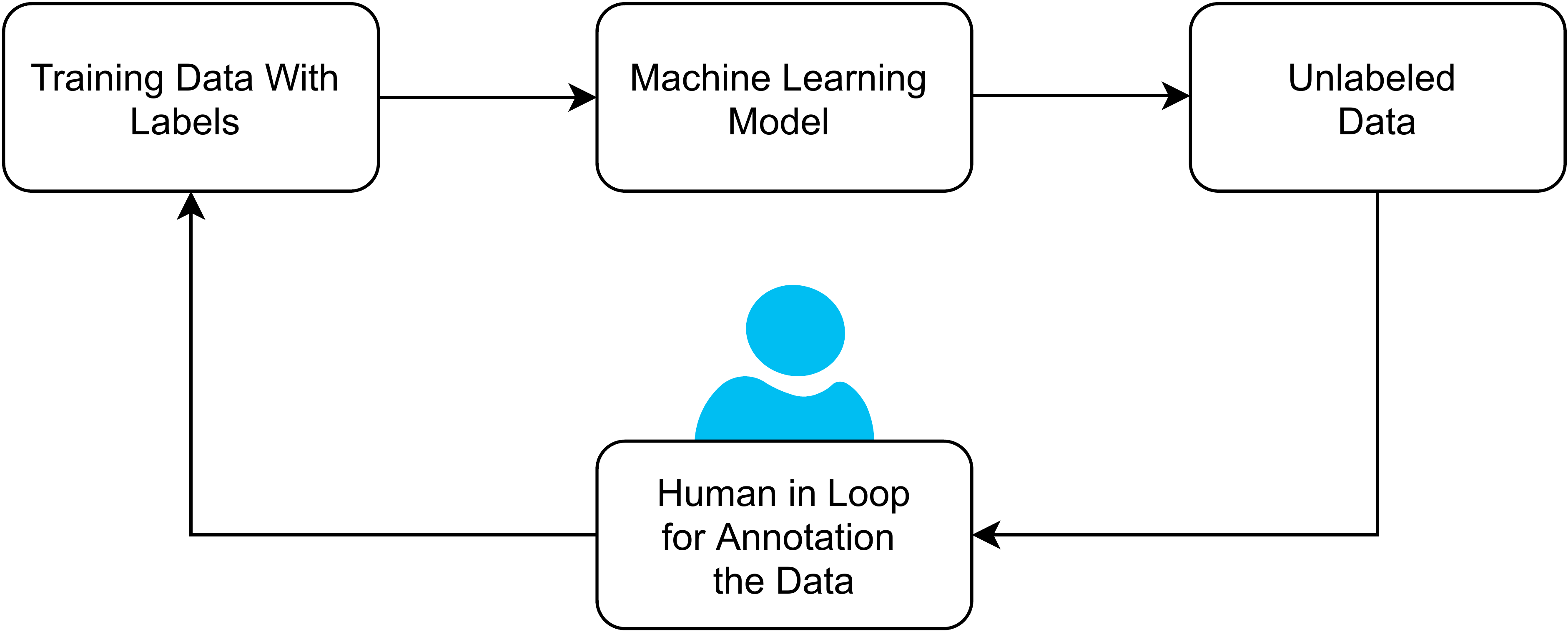}
    \caption{ Basic Framework of Active Learning~\cite{settles2009active}}
    \label{figure 7}
\end{figure}

Active Learning~\cite{thrun1995exploration} is a discipline of machine learning where the algorithm is designed for learning, can choose the data for learning, or learning strategy generated during learning (Figure \ref{figure 7}).  The active learning methodologies can play an essential role in domains that are dealing with real-time data such as speech recognition, information extraction~\cite{settles2008active}, classification, and filtering. Moreover, active learning provides high accuracy with a small testing size of labeled data.

There are different types of scenarios of active learning, such as membership query synthesis~\cite{angluin1988queries}, stream-based selective sampling~\cite{cohn1994improving}, and pool-based active learning~\cite{lewis1994sequential}.
The standard data mining methods learn models with isolated data and make a prediction based on static models~\cite{yin2006efficient,kuncheva2007classifier,baralis2007lazy}. It needs to use previous knowledge, or a learning model should transfer knowledge, and it must be used to predict future learning. It is termed as transfer learning~\cite{pan2009survey}. The knowledge can be transferred in various forms such as transferring knowledge of instances~\cite{dai2009eigentransfer}, knowledge of feature representations~\cite{argyriou2008convex} (for both supervised and unsupervised), knowledge of parameters~\cite{lawrence2004learning} and relational knowledge~\cite{mihalkova2007mapping}.

\subsection{Lifelong Learning/Continual Machine Learning}

\begin{figure}[htb]
    \centering
   \includegraphics[height = 0.5 in, width = 5.0 in]{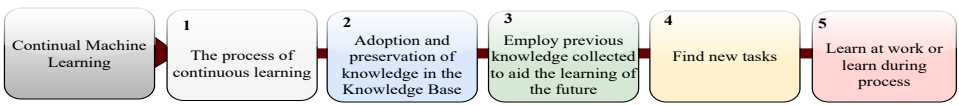}
    \caption{ Tasks of Continual Machine Learning }
    \label{Tasks}
\end{figure}

\begin{figure}[htb]
    \centering
   \includegraphics[height = 2.6 in, width = 4.7 in]{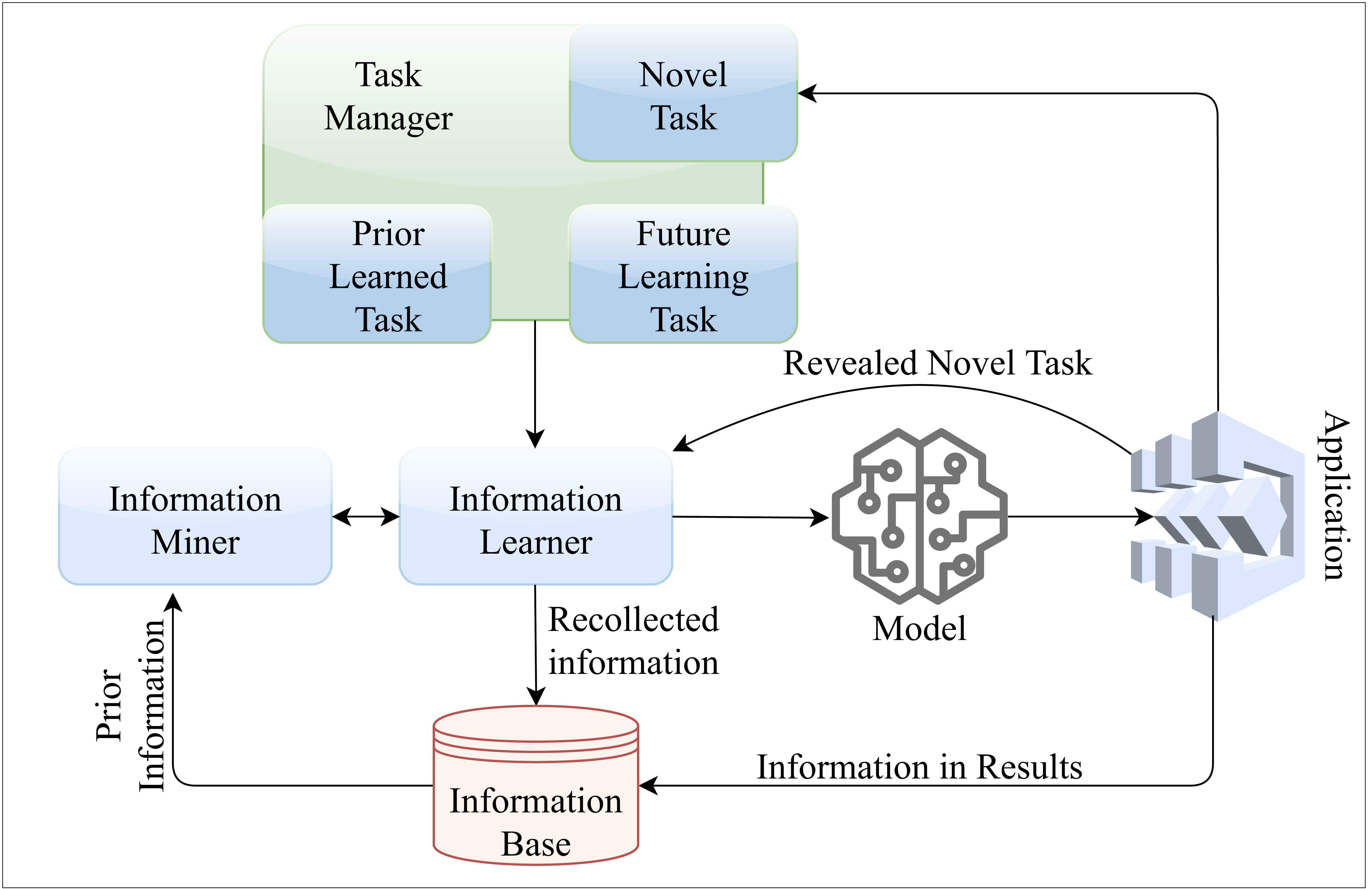}
    \caption{ Basic Framework of Lifelong Machine Learning~\cite{chen2018lifelong}}
    \label{figure 8}
\end{figure}

Lifelong machine learning is a system that can continuously learn from different domains, and this knowledge can be used effectively on future tasks in an efficient manner~\cite{silver2013lifelong}. The selective knowledge is transferred when learning a novel task. Knowledge is retained from a different source and improves learning (Figure \ref{figure 8}). The various techniques of lifelong learning as prior works in knowledge retention and improves learning a new task. The major tasks of lifelong machine learning are shown in Figure \ref{Tasks}. 

There are different names: constructive induction, incremental and continual learning, explanation-based learning, sequential task learning, and never-ending learning. These methods are further divided into different categories: lifelong machine learning is supervised learning, continual learning is reinforcement learning, and self-taught learning or never-ending learning is unsupervised learning.

Supervised learning lifelong learning uses Explanation-based Neural Network (EBNN) using back propagation gradient. Whenever new learning tasks occur, EBNN uses prior domain information of the task. EBNN gives more accurate results even with fewer amounts of data. In~\cite{shultz2001knowledge}, authors suggested knowledge-based cascade correlation neural networks. This method uses prior trained networks and concealed units to set the new bias for a novel task. Unsupervised Lifelong learning is used to increase the system's scalability, and adaptive resonance theory had been used to map the bottom and top nodes of clusters. Set threshold to consider a new example node, a map with vigilance parameter (below threshold).

In~\cite{strehl2002cluster}, authors proposed a novel approach to ensemble clusters from the primary partition of objects; it uses labels of the cluster deprived of accessing the original features.  The self-taught learning models build high-level features by using unlabeled data and test such models for various classification applications in image, web, and song genres. Lifelong learning goals can be achieved by another popular method is Never-ending Language Learner (NELL)~\cite{carlson2010toward}. NELL extracts data or reads information from the web and increases its knowledge, then learns how to perform a new task better than the same task done in the earlier day.

Rather than focus on conventional machine learning, the system should retain knowledge and transfer this to the system to the learning agent. The system should learn sequential tasks and increase their magnitude.

\subsubsection{Challenges and Benefits of Lifelong Learning Models~\cite{silver2013lifelong}.}

\begin{itemize}
\item \textbf{Input /output Type, Complexity and Cardinality :} The real-time environment has a variety of data from different domains; it can differ in nature. The attributes of each input may vary according to their source and required task.

\item \textbf{Training Examples Vs Prior Knowledge :} In life-long learning systems, prior knowledge is a crucial part of the end-to-end system to achieve accuracy while performing a new task. There is a need to retain valid data from the knowledge base that must have information act as a training example.
    
\item \textbf{Effective and Efficient knowledge Retention : } The system must retain efficient information that must not be erroneous. Furthermore, it must use finite memory to store knowledge with limited computational capacity. The system must be capable of handling duplicate data and increase the accuracy of the prior knowledge.

\item \textbf{Effective and Efficient knowledge Transfer :} Prior knowledge should not increase computational time and effort. Moreover, the transfer of knowledge shout not generates less accurate inputs/models for new tasks. There are three major components of lifelong learning.
\begin{enumerate}
     \item Retention of learned task knowledge,
     \item Selective transfer of prior while learning a new task, and
     \item The system must ensure that retention and transfer of knowledge must be efficient.
\end{enumerate}

\item \textbf{Scalability :} Scalability is one of the most challenging and essential aspects of almost all fields of computer science. The system must be able to adapt increments in volumes of input data. The lifelong learning systems must be able to address the space and time complexity of all these factors.

\item \textbf{Heterogeneous Domain of Task :} The lifelong learning systems must handle data from different domains by establishing relations among the origin domain and targeted domain. There are so many features that are common between but diversity in transferred knowledge data also exists. The system must have the ability to map features in transferred knowledge.
\end{itemize}

\subsection{Multi-task Learning}

Multi-task learning  (MTL)~\cite{caruana1997multitask,chen2009convex,li2009multi} acquires various associated tasks concurrently, beaming at delivering a more reliable representation by using the associated knowledge yielded by various jobs.  The motive behind introducing inductive bias in Multi-task learning is to joint hypothesis space of every job by utilizing the task-relatedness building. It additionally inhibits over-fitting in the specific job and therefore has a more immeasurable generalization capability. Unlike transfer learning, it mainly does various jobs preferably than various areas as much of the area's existing research is based on several comparable jobs of the identical application area. Multi-task learning allows those jobs are strictly associated with each other. There are several hypotheses in terms of job-relatedness, which drive to another modeling strategy. 
Many researchers continue to hypothesize that all job data come from the same sources and are correlated to the standard or global models. According to this hypothesis, they created the association among jobs employing a task-coupling parameter, including regularization.
In~\cite{liu2015representation}, authors proposed multi-task learning for the deep neural network. They classify multi-tasking tasks into two categories for deep learning. First is classification, and second is ranking; in classification, the model identifies the queried domain, whereas the ranking model finds the relevant queries.

\section{Research Challenges}
Learning in a dynamic environment is still a challenging task due to the unpredicted nature of the upcoming event. How we can integrate the classifier to obtain a sub-knowledge of unknown classes and reduce the open-space risk. The significant challenges in open-world machine learning  are:
\begin{itemize}
    \item \textbf{Incremental Volume of Data:} This is the era of digitization. Hence various sources are generating a large amount of data. This data is not only significant in volume but also unstructured. Managing and finding the different classes of the various domains is very difficult as continuous updates appear with additional unseen instance categories. There is a lack of a mechanism that can deal with real worlds data.
    
    \item \textbf{Identifying a Novel Classes:}	Once a system identifies instances as unseen or rejects unseen classes, the system has to learn about the classes of these unseen instances. There is a need for a complete framework that can address these unseen instances and make novel classes for them. The framework must find the instances of each class to acquire new classes.
    
    \item \textbf{Updating a Knowledge-Base:} There is a lack of a mechanism that can append new knowledge to the system at run time. There are many complexities in appending a new domain and its classes in a progressive environment as input data overgrows. There is also a need to use such classes (newly recognized) for the next prediction without retraining the entire model.

    \item \textbf{Open-space Risk:} During the learning phase in the open environment of the open world, it is hard to manage space away from positive training examples. It is referred to as open-space. The open-space risk needs to be addressed to learn more accurately with increasing openness. There is a need to build a framework that can reduce open-space risk.

    \item \textbf{Open Framework:} There is a need for a  generic framework to discover unseen classes in the real-world domain that can function end-to-end for learning in a progressive environment. To build the complete framework of an open-world machine learning  system, one needs to execute both operations together ( discovery of unseen instances and identification of novel classes).  Hence it needs two or more different modules to function dynamically. These modules can use different methods; Hence modules' concatenation and synchronization are relatively complicated as different methods are involved. The model needs to precisely address both seen and unseen classes with synchronization of newly adopted classes.

    \item \textbf{Efficient Discovery of Novel Classes:} There is a lack of an open-world machine learning  model to discover and identify the unseen classes. Various models exist in open-world machine learning  to discover unseen instances, but very few can identify the novel classes out of these unseen instances. However, there is also a lack of learning models that identify the number of unseen classes efficiently.
    
    \item \textbf{ Retention of Obtained Knowledge:} There is a lack of models that continuously updates the knowledge while learning a new task. To the best of our knowledge, very few learning models use a continual learning approach to learn unseen instances employing previously obtained knowledge for the subsequent prediction.
    
\end{itemize}

\section{Future Directions}
We have reviewed numerous research works of open-world machine learning  in computer vision and image processing, and natural language processing. Based on the study, we have identified three significant aspects necessary to achieve learning in the open-world. open-world machine learning  can be improved by enhancing these three aspects, model, rejection capability, and identification of new classes. 
This section discusses and analyses the limitation in brief and discusses the research directions in detail.\\

\noindent\textbf{Open-world Models.}
The existing models of open-world machine learning  are working in a hybrid manner and address the problem in parts. There is a lack of models available that can work in an end-to-end manner. The end-to-end model for open-world machine learning  can strengthen the classification for both categories, known-known class, and known unknown class. To our best knowledge, there is no promising model available for unknown-unknown class identification, which is one of the challenging categories in open-world machine learning . The existing methods for unknown-unknown class classification are worth extending further.

\noindent\textbf{Rejection of Unknown Classes.}
Few works are available to reject unknown-unknown classes, while automation systems entirely depend on unseen class rejection with high accuracy. Further work to increase the rejection capability of unknown classes can make the system more reliable as the real-world application faces many unknown objects while working in a dynamic environment. Existing models need more improvements to reject unseen classes with high accuracy.

\noindent\textbf{Identification of Unseen Classes.}
Most existing models either detect known or reject unknown, but after rejecting classes as unknown, there is no promising mechanism available to further identify classes in rejected data. There is a need for models that can identify the number of hidden clauses in rejected data.

\section{Conclusion}

In this paper, we investigated the works in open-world machine learning proposed in the last decade. We have also discussed the prominence and many real-life applications of open-word machine learning. Many algorithms, models, and frameworks have been proposed in the literature to address numerous objectives allied to open-world settings. The domain is relatively new; thus, there are inadequate sources of information. The presented review will help in understanding the open-world scenario, working, and associated challenges. We provided a task-based classification of OWML in CV-IP and NLP.   Further, we discussed various techniques and used datasets in OWML.  The limitations of numerous technologies are also analyzed to facilitate promising future extensions of these methods.

\section*{Declaration of Competing Interest}
The authors declare that they have no known competing financial interests or personal relationships that could have appeared to influence the work reported in this paper.

\bibliographystyle{ACM-Reference-Format}
\bibliography{main}

\appendix

\end{document}